%% file: main.tex
\documentclass[runningheads]{llncs}

\usepackage{eccv}

\usepackage{eccvabbrv}

\usepackage{graphicx}
\usepackage{booktabs}
\usepackage{multirow}
\usepackage{makecell}
\usepackage[export]{adjustbox}
\usepackage{wrapfig}

\usepackage[accsupp]{axessibility}  %

\usepackage[pagebackref,breaklinks,colorlinks,citecolor=eccvblue]{hyperref}

\usepackage{orcidlink}

\begin{document}

\title{Template-based Object Detection Using a Foundation Model}

\author{Valentin Braeutigam\inst{1}\orcidlink{0009-0006-4362-1300} \and
Matthias Stock\inst{2} \and
Bernhard Egger\inst{1}\orcidlink{0000-0002-4736-2397}}

\authorrunning{V.~Braeutigam et al.}

\institute{Friedrich-Alexander-Universität Erlangen-Nürnberg, 91058 Erlangen, Germany 
 \and
e.solutions GmbH, 91058 Erlangen, Germany \\
\email{valentin.braeutigam@fau.de} }

\maketitle

\begin{abstract}
Most currently used object detection methods are learning-based, and can detect objects under varying appearances. 
Those models require training and a training dataset. 
We focus on use cases with less data variation, but the requirement of being free of generation of training data and training. 
Such a setup is for example desired in automatic testing of graphical interfaces during software development, especially for continuous integration testing.
In our approach, we use segments from segmentation foundation models and combine them with a simple feature-based classification method.
This saves time and cost when changing the object to be searched or its design, as nothing has to be retrained and no dataset has to be created. 
We evaluate our method on the task of detecting and classifying icons in navigation maps, which is used to simplify and automate the testing of user interfaces in automotive industry. 
Our methods achieve results almost on par with learning-based object detection methods like YOLO, without the need for training. 
\keywords{UI testing, Foundational Models, Object Detection}
\end{abstract}
\begin{figure*}
    \centering
    \includegraphics[width=0.9\linewidth]{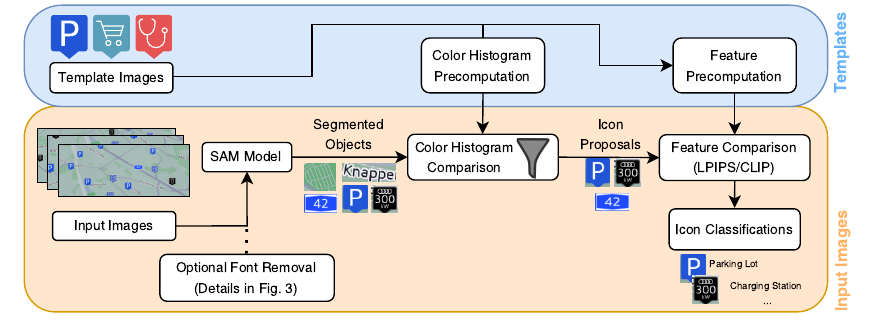}
    \caption{\textbf{Overview of Our Workflow.} We use a SAM model to segment all objects in the input image and use each segment as an object proposal. The object proposals can either be generated from the input images or the images with removed text which are generated as shown in \cref{fig:text_removal}. To reduce the number of icon proposals we first compute the correlation of the color histograms with each histogram of the templates and reject candidates without matching templates. For the remaining templates we compare the metric values (LPIPS or CLIP) between the candidate and the template crops. If there are values below a certain threshold the template with the best metric value is predicted as match.}
    \label{fig:workflow}
\end{figure*}

\section{Introduction}
In the field of object detection, template matching was considered state-of-the-art for a long time.
This has changed for most applications with the use of strong features and machine learning. 
One advantage of learning-based methods is that they can be very robust to appearance changes, but this is not required for every application.
We focus on one problem field where these methods are applied, namely in automated UI testing in the automotive context, especially for detecting icons in an automotive navigation map. 
In this use case, there is only a restricted variety of appearance changes, as the icons are changing in size, but are not rendered in perspective, and have no illumination. 
Nevertheless, there are frequent changes in the design or icons to search, while most of the tests are still implemented by manually created test cases using pixel comparisons of the renderings to template images created before. 
This is very time-consuming during creation of test cases and it requires adaptation of the test cases every time the design changes, which again takes much effort. 

One approach to be more data invariant is the use of template matching, which seems suitable to detect UI elements as there is not much variation compared to objects rendered in perspective. 
As they are orthographic this leads to the simplification that they are not rendered from different perspectives, which reduces the number of required templates. 
However, in the case of icons the problem of different scales remains, which causes problems when applying standard template matching. 
Additionally, there is the challenge of similar or partly equal icons, e.g. when having the icons for a parking lot and a parking garage. 
A third problem can be other objects that partially cover the searched objects.
In our example, this is the case with text, e.g. city names or street names, that is rendered in front of the icons. 

One solution to deal with all of these problems, which is already in use for automated testing~\cite{branco2023deep} are learning-based methods. 
There are various learning-based object detection methods, that have shown a good accuracy~\cite{redmon2016lookonceunifiedrealtime,rcnn2014,liu2016ssd,zhang2018single}, and some methods also while predicting results in real time, e.g. the YOLO models~\cite{redmon2016lookonceunifiedrealtime}. 
With these methods the presence can be checked automatically, based on classes specified by the training data, and the models can be trained to perform on different scales of searched objects. 
However, in this scenario changing designs remain a problem, as the methods have to be newly trained to adapt to the new unseen objects. 
This is especially a problem, when using continuous integration, as the test pipeline would have to be updated, re-trained and checked every time the design changes.
Another disadvantage is the requirement of a training dataset, which has to be created and newly generated or adapted with each design change. 
Additionally, the dataset has to cover the possible positions of the icons on the different possible backgrounds and icon sizes. 

Especially during a design phase where the appearance of tested items often changes, this can be time consuming and is therefore not suitable for integrated testing. 
We present a new method, which is a modern intersection between template matching and object detection methods that combines advantages of both.
Although it is not restricted to the depicted use case, we evaluate it for this one. 
The method uses a foundation model for segmentation, concretely Segment Anything 2.1 (SAM2.1)~\cite{ravi2024sam2segmentimages}, respectively SAM3~\cite{carion2025sam3segmentconcepts}, in combination with a template-based classification. 
For the second part features generated by publicly available pre-trained networks from commonly used methods like CLIP~\cite{radford2021learning_clip} or LPIPS~\cite{lpips} are used. 
With this approach, we circumvent the need to re-train and collect datasets for the training while still being scale invariant and robust to partial occlusions, which combines the advantages of both approaches. 
In addition, we extend our approach by segmenting text in the input images that potentially covers the searched elements. 
Afterwards, we apply inpainting to the image masks to remove the text in the image and to improve classification results.
Our code is publicly available at \url{https://github.com/valentinBraeutigam/Template-Based-Object-Detection}. 
\\
Our contributions are as follows:
\begin{itemize}
    \item we present a method to detect and classify objects in an image with no need for learning/adapting to new objects
    \item instead of a dataset we need only one template per object, each of them can be easily replaced or modified
    \item our method is scale-invariant and robust to partial occlusion by text removal
    \item we provide source code of the proposed template-matching framework
\end{itemize}

\section{Related Work}

\noindent\textbf{Template Matching.} There are several template matching techniques capable of detecting a given template in an image~\cite{brunelli2009template, zhang2021low,liu2017fast,kim2007grayscale,yang2019large}. 
These algorithms are usually based on a sliding-window approach~\cite{brunelli2009template} which means that they iterate over the whole image and compare the template at each window position with the input image by a metric.
Commonly used metrics are squared difference, cross-correlation, and correlation coefficient. 
These techniques have the problem that they are often not size invariant, which is usually solved by iterating over the image with different scaling factors of the template image, e.g. in ~\cite{zhang2021low}. 
But this can be very tedious as, depending on the size changes that shall be covered, the process has to be repeated multiple times.
Apart from the runtime, it requires the user to precisely tune a threshold to differ between objects and background, and the step size depending on the problem to not miss a good match between two different possible sizes, which can be sensitive when having small features. 
Those methods were very popular before the rise of strong features and deep learning, current detection algorithms focus on strongly varying object appearance and are more robust to changes in the appearance. 
For objects or icons with constant appearance without a background containing strong features, there is little literature, and we did not find a combination of classical template matching with state-of-the-art image features and segmentation models.

\noindent\textbf{Feature Matching.}
Feature matching methods compute image features from both a template image and an input image. 
Then, the image features are compared to find matching pairs of features between both images. 
In that regard, multiple feature detectors can be used and combined with various matching algorithms.
A prominent method is the SIFT (Scale-Invariant Feature Transform) algorithm by Lowe et al.~\cite{lowe2004distinctive} to compute image features and the FLANN (Fast Library for Approximate Nearest Neighbors) algorithms~\cite{muja2009fast} to detect matches across the images. 
These approaches are capable of working scale-invariant, but they are designed for finding matching features instead of all occurrences of an icon.
Therefore, a threshold would be needed to define how many corresponding features have to be found to define a match, and then the template could be fitted to the image by scaling it according to the found features. 
In addition, feature matching is probably better suited for photorealistic images than for finding icons in a synthetic environment, as they usually contain more unique features. 
Photorealistic images often have more unique features compared to icons that consist of sharp boundaries and sometimes share the same image features with other icon classes, as they can consist of equal or similar parts. 

\noindent\textbf{Object Detection.} Object detection is a well-known problem in computer vision and since there exist many methods achieving a good performance on photorealistic in-the-wild images, they also are suitable to find objects in images with a synthetic appearance. 
In fact, this seems to be the easier challenge due to the missing lighting, reduced texture and the fact that UI elements, which are the searched objects, are not rendered in perspective. 
Some well-known object detection methods are the YOLO models~\cite{redmon2016lookonceunifiedrealtime, Ultralytics_YOLO,Ultralytics_YOLO_documentation} or the R-CNN models~\cite{rcnn2014}. 
In comparison to our approach, they have the main disadvantage that they need to be trained on images containing the searched templates. 
Whilst the resulting methods are highly robust to differences in appearance, this is not required for some tasks, for example, our application of automated testing of UI elements.
That also includes the need for a training dataset containing the searched icons in different scales on the different possible backgrounds, which needs to be created before training and testing can start. 
This already requires a correct working renderer, which shall be tested with the method. 
In our experiments, we compare our method with two of the YOLO models for comparison of accuracy. 

\noindent\textbf{Instance Segmentation.} There are several approaches in the field of instance segmentation of unseen classes from rgb images, e.g. Du et al.~\cite{du2021learning} or Zhang et al.~\cite{zhao2022novel}. This is a related field, but differs as our approach aims to detect known objects. 
Closest to our work, Nguyen et al.~\cite{CNOS_nguyen_2023} present an approach for detecting and segmenting objects in rgb images. To achieve this, they compute DINOv2~\cite{oquab2024dinov2learningrobustvisual} features for images of 3D objects rendered from different view points. Then, they compare the features to segments generated with SAM from input rgb images, to find the best fitting 3D object. Whilst the representation and classification are implemented similarly, the feature extraction differs in the 2D compared to the 3D setting. 

\noindent\textbf{Automated Testing of User Interfaces.}
In the past, there were several approaches to automate the testing of user interfaces (UIs) especially in the field of automotive infotainment systems. 
Huang et al.~\cite{huang2010development} present a hardware setup for automated testing of an infotainment system. 
They utilize a camera setup to capture the display output while simulating user input via hardware and then check if the visual outputs recorded by the camera match the specifications.
This enables automated testing in the sense that they do not have to be done manually by a human, but the design of the test cases is still time consuming. 

Branco et al.~\cite{branco2023deep} show the use of deep learning methods in automotive UI testing. 
They compare different convolutional neural networks (CNNs) that are commonly used for object detection for the detection and classification of UI elements in an infotainment system. 
Their approach involves creating automated tests from recorded manual tests.

In a more general context, there are also various directions of research that tackle the problem of automated UI testing. 
Yang et al.~\cite{yang2013grey} present an approach to create a model of possible GUI states and events that can be systematically traversed afterwards.
Pezz\`{e} et al.~\cite{pezze2018automatic} compare different techniques to traverse GUIs to reach optimal test coverage.
Nass et al.~\cite{nass2021many} give an overview of existing problems, giving recommendations for future research.
In our work, the aim is to simplify testing by abstracting the image detection part from the actual validation of the rendering.  %

\noindent\textbf{Object Removal and Image Inpainting.} 
There are several methods that deal with watermark removal, which is close to our task of removing obstacles in the form of text.
Most of them use a neural network architecture to predict a mask of the watermark and then inpaint the region~\cite{Niu_2023_ICCV, Liu_2021_WACV,  Chen_REFIT_2021, fu_watermark_removal_uNet, Tain_CNN_Watermark_removal_2024}.
There exist also methods explicitly used to remove text from images~\cite{Pnevmatikakis_Inpainting_system_automatic_image_structure_2008, Wagh_Text_detection_removal_2015, Liu_EraseNet_2020, Khodadadi_Text_localization_inpainting_2012, modha2014image}. 
Many of these methods are designed for axes-aligned text and with a photorealistic background. 
Liu et al.~\cite{Liu_EraseNet_2020} use an end-to-end learning approach to train a GAN-based model to remove text with any orientation and curvature from natural images.
However, there is a domain gap between natural images and synthetic image, as in our case. 
\\
In order to also handle text, independent of orientation and curvature, and without training a model, we use the color consistency of the font color across the images in our data.

\section{Methods}
\label{sec:methods}
\begin{figure}[h!]
    \centering
    \includegraphics[width=0.65\linewidth]{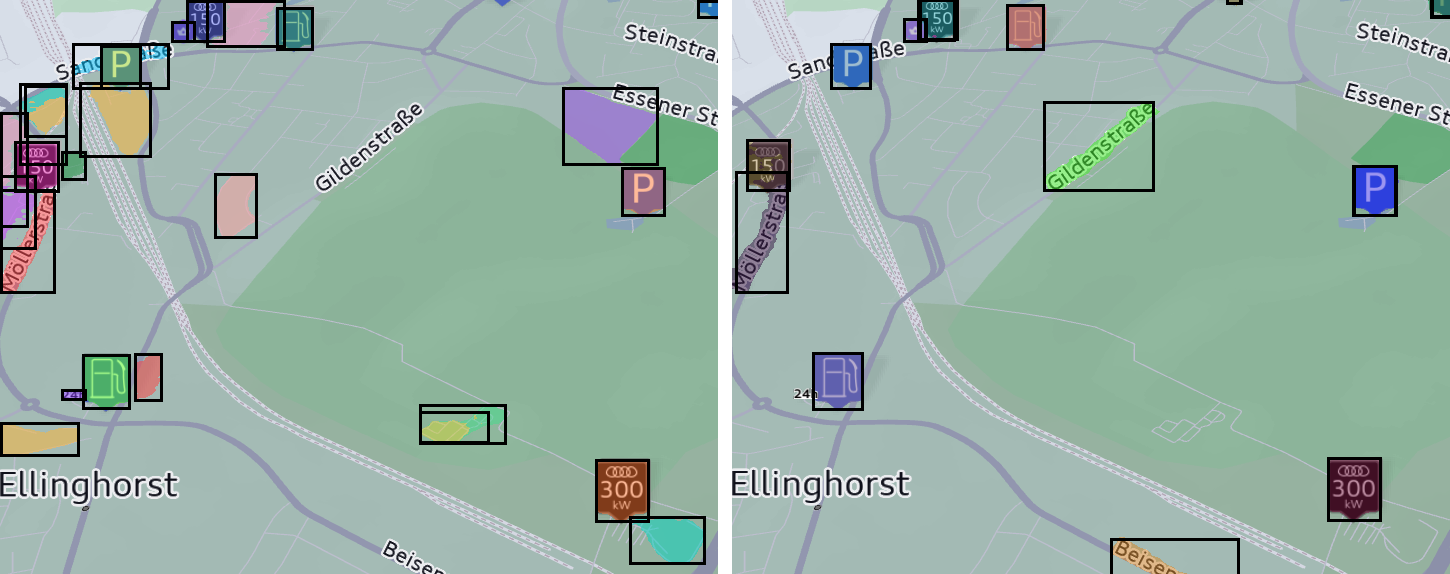}
    \caption{\textbf{Example SAM Segment Images.} %
    The segmentation masks are colored and surrounded by a black bounding box for visualization.
    This includes icons, text, and also parts of the background. 
    These proposals will afterwards be filtered during the classification process. The left image shows the result from SAM2.1, the right one the result from SAM3.}
    \label{fig:sam_detection}
\end{figure}
\label{sec:remove_text}
Our method consists of several steps, that are summarized in \cref{fig:workflow}. 
They include the segmentation of the input with SAM, text removal, feature extraction and classification. 

\noindent\textbf{SAM Image Processing.} 
The input images are processed by segmenting each image. 
As SAM segments objects based on a given input prompt, such as a point or bounding box, we segment the whole image by creating a point grid as input prompt and then using SAM2~\cite{ravi2024sam2segmentimages} to get the best segment mask for each point. 
The number of points can be adjusted according to the estimated size of searched objects to lower computational costs. 
A preselection of the proposals is done by SAM2.1 by filtering segments with low confidence or with very small, respectively large, segment masks. 
In order to do so, the segments are filtered by different properties inside the automatic segmentation annotator provided with the SAM2.1 model. 
These properties are defined in the SAM2.1 implementation~\cite{SAM2.1_github}. 
The first one is a threshold for the predicted intersection over union (IoU) between predictions and ground truth segments from the model. 
The second one is a threshold for the stability score which uses the stability of the masks while changing the cutoff used to binarize the masks. 
These properties can be lowered to increase the number of proposed segments. 
SAM3 introduces the capability to search for concepts by providing images or textual categories as input prompts. 
Similar to SAM2 it has a confidence threshold to filter predictions with low confidence. 

\begin{figure}
    \centering
    \includegraphics[width=0.7\linewidth]{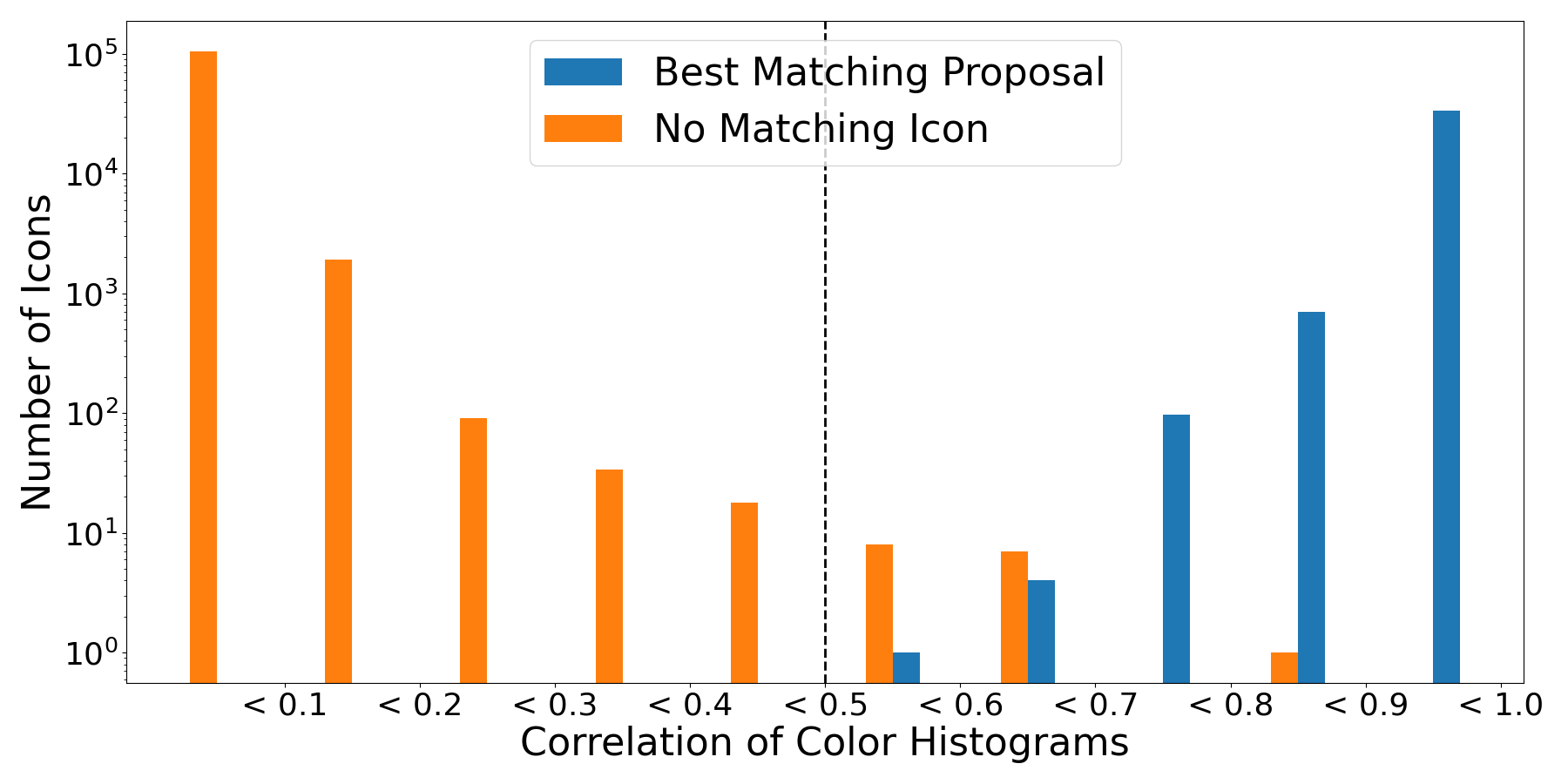}
    \caption{\textbf{Comparisons Between the Color Histogram Correlation of Icons and Non-icons}. The best correlations for proposals without complete icons, and the correlation values for each icon with the best matching proposal are counted. The correlations are grouped into bins with a width of 0.1 and the number of icons are scaled logarithmically. As none of the best matches to icons has a correlation below 0.5, this part can be filtered by applying a correlation threshold (dotted line), reducing the number of icon proposals. } 
    \label{fig:correlation_plot}
\end{figure}
Based on the segment masks, we compute the bounding box of the detected proposals by using the minimal and maximal coordinates along each axis of the segment mask. 

\noindent\textbf{Feature Extraction and Classification.} 
As the icon proposals computed by SAM also contain crops only covering background, we first filter the proposals by their color.
To achieve that, the correlation between the color histograms of the proposals and templates is computed to check for similarities in the color distribution. 
If the highest correlation is below a set threshold, the object proposal is withdrawn.
This is usually the case when there is no icon in a proposal or the proposal is degenerated, meaning that it is only a small part of an icon or an icon combined with a large portion of background.
Degenerated proposals are usually not the only proposals containing that icon, as the icons have clear boundaries, which leads to a good segment of the icon by SAM in most cases. 

\Cref{fig:correlation_plot} shows a plot with the number of icon proposals per correlation value. 
It is grouped into icons with their best matching proposal, and icon proposals that do not correspond to an icon. 
One can see that icons having a match tend to have a high correlation value with the matching template, and there are no icons having a best correlation value below a threshold of 0.5. 
Therefore, many incorrect icon proposals can be removed while all correct icon proposals are kept. This leads to less time consumption, small accuracy improvements, and to less parameter tuning for the used metric. Additional visualizations are given in the Supplementary Materials.  

After comparing the color histogram correlation, we utilize similarity metrics based on computed features to classify the templates. 
We evaluate our approach with two different metrics: the cosine similarity between the CLIP~\cite{radford2021learning_clip} features and the LPIPS~\cite{lpips}. 
The features of the object candidates are compared in both cases to the pre-computed features of the templates that fulfill the color criteria. 
Templates that have a correlation below 90\% of the highest correlation between an template to the current proposal are withdrawn, to speed up performance and to ensure that the color is taken into account when icons have a very similar pattern but a different color.

For both models, the object crops are preprocessed by rescaling them to an input shape of 224$\times$224 and normalizing them based on the ImageNet mean and variance.
For LPIPS, there exist two pre-trained variants, one with a VGG model and one with an AlexNet model as basis.
In our work, we use the AlexNet variant, which the authors state has a better forward score, while the VGG variant is closer to 'traditional' perceptual loss~\cite{lpips}.
For the CLIP features, we found that we achieve better results when scaling them according to the minimum and maximum values for each feature gathered from the template data. 

In a last step, we apply non-maximum suppression by filtering the remaining icon proposals for bounding boxes with an IoU with more than 10\% and using only the bounding box with the best metric result. 

\noindent\textbf{Removing Font.} 
\begin{figure}
    \centering
    \includegraphics[width=0.9\linewidth]{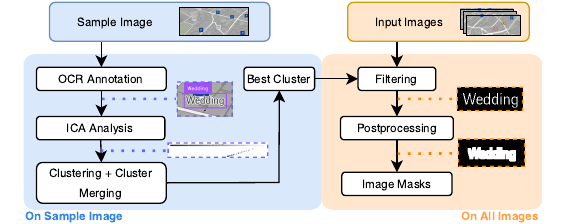}
    \caption{\textbf{Overview of the Font Removal Method.} 
    We start computing color clusters from optical character recognition (OCR) bounding boxes of a sample image. 
    Then, we apply an independent component analysis (ICA) and cluster the projected colors. %
    Similar clusters are then merged, and the best fitting cluster is selected by comparing the area filled with colors from each cluster inside and outside of OCR bounding boxes. %
    Given the best cluster, we create image masks for each input image by filtering by the best color cluster and post-process by dilation and erosion.
    These image masks are afterwards used to inpaint the parts containing text before classification. }
    \label{fig:text_removal}
\end{figure}
To optimize the accuracy of our method, we additionally remove text, e.g. city or road names, from the images, which covers icons we aim to detect. 
We use the assumption that the font consists of the same color cluster throughout the images, which is potentially interpolated with other colors during rendering. 
An overview of the method is given in \cref{fig:text_removal}. 

First, we compute bounding boxes by optical character recognition (OCR). 
Next, we identify color clusters from that part of the image by applying an independent component analysis (ICA) and clustering the projected colors using the BIRCH clustering algorithm by Zhan et al.~\cite{birch_zhang_96}. 
We reduce the color clusters by comparing the color clusters across a few samples in the image and removing colors not shared across the samples.
In a last step, we select the best matching cluster by the ratio between masked pixels detected inside of the OCR bounding boxes of the sample image compared to masked pixels in total multiplied with the number of detected OCR bounding boxes. 

With the resulting font color cluster, we compute segment masks by filtering each image according to the cluster.
In a postprocessing step, the masks are dilated and eroded to remove small, unconnected pixel groups and to remove the outline of the text.
To the areas covered by the text segment masks, we apply an inpainting model to replace the parts with text in the images. 
We do not expect to reconstruct the complete icon as the model is not finetuned on the navigation map design, but we expect that perceptible features of the text are reduced leading to fewer misclassifications. \\

\section{Experiments \& Results}
\label{sec:experiments}
We evaluate our method against the object detection method YOLO.
Additionally, some experiments are done with template matching and feature matching algorithms using existing implementations.
However, they are not suitable for evaluation of the entire datasets, as we could not observe convincing performance on evaluations on a small amount of data.
For our evaluations, only icons that are completely in the image are used. 
Icons that have pixels at the image border and are therefore expected to have pixels outside the image are ignored. 
Predicted icons are mapped to the ground truth annotation by matching bounding boxes where the midpoints are positioned in the other bounding box. 

\begin{figure}
    \centering
    \begin{subfigure}{0.49\textwidth}
    \includegraphics[width=\linewidth]{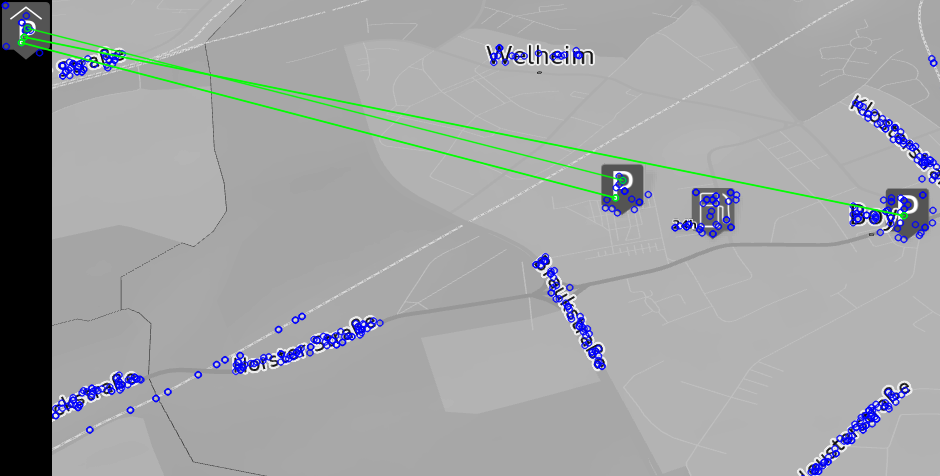}
    \caption{\textbf{Feature Matching.} The blue circles highlight detected SIFT features in the template and input image. The green lines show matching features between the searched template in the left and the image. There are matching features found in two icons from a different class. }
    \label{fig:results_feature_matching}
\end{subfigure}
   \begin{subfigure}{0.49\textwidth}
    \includegraphics[width=\linewidth]{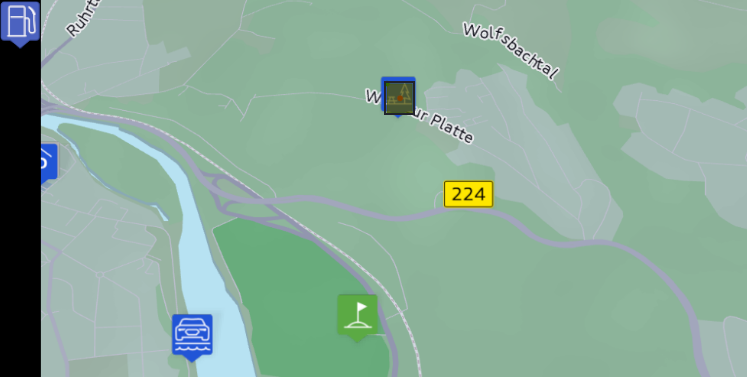}
    \caption{\textbf{Template Matching.} The searched icon is the gas station icon displayed in the left of the image. However, it is mismatched with the icon of a picnic place. Especially for icons that are not contained in the image, the metric (in this case squared differences) often finds a false positive. }
    \label{fig:results_template_matching}
\end{subfigure}
\caption{\textbf{Qualitative Result of Template and Feature Matching Algorithms.} Some qualitative results to template and feature matching approaches showcasing the problems when applying the methods to the given use case. }
\end{figure}
\noindent\textbf{Feature Matching.} 
To explore feature matching algorithms, we used SIFT features followed by either a brute-force matching or the FLANN-based knn-matcher on the features. 
However, for most icons, none or only a small number of features are found in the image, which is not sufficient for matching the icon to a place in the image. 
Especially, as there is a similar amount of matches with the background or icons from a different class. 
One case where this wrong detection often occurs is when icons share parts with equal design. 
An example with the parking lot and parking garage icon is shown in ~\cref{fig:results_feature_matching}.
As we do not find consistent matches with the icon to classify, we do not evaluate the experiments quantitatively. 

\noindent\textbf{Template Matching.}
For template matching, we experiment with an implementation by Zhang et al.~\cite{zhang2021low} that uses different scales of the searched templates to find them in the provided image. 
We use three different metrics (cross-correlation, correlation coefficient, and squared difference) for comparison of similarity to the searched icon. 
A qualitative result is shown in \cref{fig:results_template_matching}. More results can be found in the Supplementary Material.
However, the number of correctly classified icons in a small set of images was low, while having a high number of false positives. Additionally, template matching requires individual tuning of the threshold parameter. 

\noindent\textbf{Datasets.}
We evaluated our approach on two datasets of navigation system renderings, provided by the company e.Solutions GmbH, each of which has individual designs for backgrounds and icons. 
The datasets are created with a map renderer with the original design used by two different car manufacturers. 
Additionally, the icon templates are given in SVG format, which are rasterized to PNG format and then used for matching the icons. 
The datasets consist of 15,855 (dataset \textit{A}) and 37,260 images (dataset \textit{B}), with a size of 1800 $\times$ 697 pixels. 
Both datasets contain 85 classes of icons.
Although our method does not require training, we need training data for the evaluation of the YOLO models, and therefore we split it by the ratio of 70\%/30\% into training and testing set.
We evaluate both methods on the testing sets, which contain 4,741 images (34,122 icons) for dataset \textit{A}, and 11,168 images (64,436 icons) for dataset \textit{B}.

\noindent\textbf{Image Segmentation.}
\begin{table}[h]
  \caption{\textbf{Percentage of Icons Not Detected by the Segmentation Model}. Some icons are not detected as objects by SAM, for example, if the icon is covered by much text, or if something close is predicted to be more certainly an object. The quality of the segments is given by the percentage of not detected icons and the mean intersection of the bounding box of the segment with the ground truth bounding box. The SAM2.1 models are evaluated with an equally distributed point grid having 64 points along the longer image dimension. The models are evaluated on dataset \textit{A}.} %
\fontsize{8pt}{8pt}\selectfont
  \centering
  \begin{tabular}[width=0.9\linewidth]{c c c c c c}
    \toprule
    \textbf{Model} & \textbf{ Std. Thresh. } & \textbf{ Icons w/o Seg. Mask $\downarrow$ } & \textbf{ \#Predictions } & \textbf{ $\bar t_{img}$ $\downarrow$} \\
    \toprule
    FastSAM small & no & 170 / 34,122 (0.498\%) & 339,302 & 92.3 \\ %
    FastSAM large & no & 10 / 34,122 (0.029\%) & 230,944 & 376.5 \\ %
    \midrule
    SAM2.1 small & yes & 1593 / 34,122 (4.669\%) & 71,591 & 2,699.2 \\ %
    SAM2.1 large & yes & 884 / 34,122 (2.591\%) & 85,613 & 2,863.2 \\ %
    \midrule
    SAM2.1 small & no & 43 / 34,122 (0.126\%) & 235,718 & 3,947.0 \\ %
    SAM2.1 large & no & 5 / 34,122 (0.015\%) & 226,061 & 4,007.6 \\ %
    \midrule
    SAM3 & no & 1 / 34,122 (0.003\%) & 57,553 & 640.8\\

    \bottomrule
  \end{tabular}
  \label{tab:sam_segmentation_comparison}
\end{table}
To find the best segments for our use-case, we compare the SAM2.1 model, using the small and large architecture with the pre-trained weights provided by facebook research~\cite{SAM2.1_github}, the SAM3 model by facebook research~\cite{carion2025sam3segmentconcepts} and the small and large FastSAM models by Zhao et al.~\cite{zhao2023fast}. 
The results for these networks using multiple configurations are shown in ~\cref{tab:sam_segmentation_comparison}.
For the SAM2.1 models, we experimented with different numbers of points per side.
The number of points defines the longer side of the uniform point grid used as an input prompt to the SAM2.1 models. 
Along the other side the number of points is divided by the ratio between the sides, to not to increase the number of input prompts unnecessarily. 
When using SAM3 we give "icon" as an textual prompt and one icon as an image prompt to segment all the icons in the image. 
The other variation are the thresholds used.
For FastSAM, we reduced the IoU threshold from 0.5 to 0.3 when differing from the standard thresholds. 
For SAM2.1, we changed the two thresholds (predicted IoU threshold and stability score threshold) from (0.88, 0.95) to (0.5, 0.7).
For SAM3 we use a confidence threshold of 0.3. 
We compare the average time consumed per image and the number of ground truth icons that are not part of the SAM segments. 
To evaluate the accuracy of our approach, we use the results of the SAM2.1 large model with 64 points on the longer side and lowered thresholds as well as the SAM3 model, as they recognize the largest number of icons from the images. 

\noindent\textbf{Hardware.} For training of the comparison methods and evaluation, we use a workstation with an AMD Ryzen 9 7950X processor having 32 cores and an NVIDIA GeForce RTX 4060 Ti graphics card. 

\noindent\textbf{Inpainting.}
\begin{figure}
    \centering
    \includegraphics[width=0.7\linewidth]{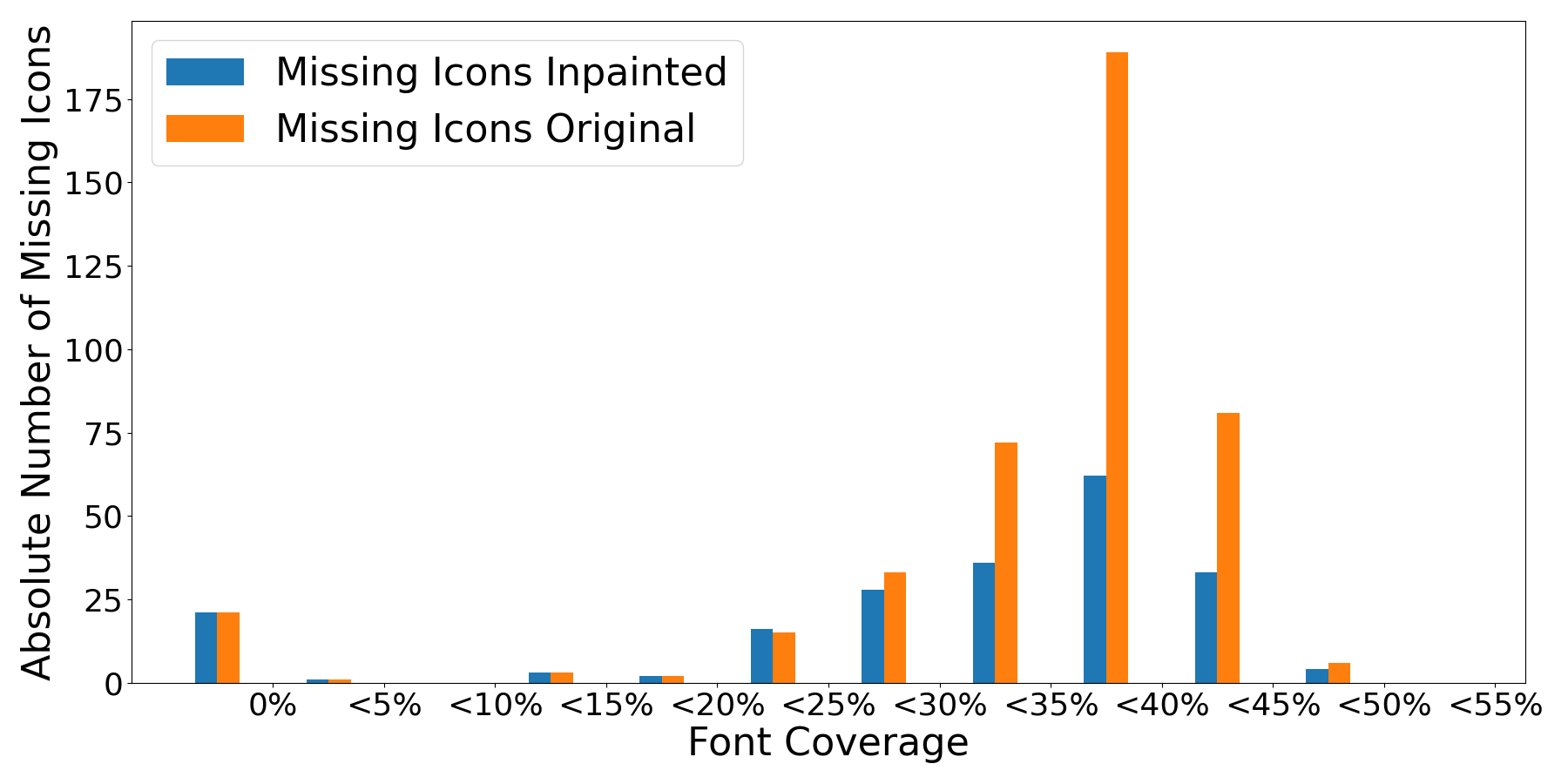}
    \caption{\textbf{Comparison Between Text Coverage and Detection Rate.} The ground truth bounding boxes are grouped by the ratio of text coverage. 
    The total number of by the classification undetected icons per text coverage bin is displayed for the method using LPIPS with and without inpainting. } %
    \label{fig:inpainting_plot}
\end{figure}
For the optional inpainting, we use the Inpaint Anything model by Yu et al.~\cite{yu2023inpaint}. 
We compute text masks and apply inpainting for each image as described in ~\cref{sec:remove_text}. 
An example for an inpainted image is given in the Supplementary Material. 
In \cref{fig:inpainting_plot}, we show the percentage of wrongly detected icons grouped by the coverage of the corresponding ground truth bounding boxes with text.
One can see that inpainting helps for a high text coverage, while having an equal performance for a low coverage. \\

\noindent\textbf{Time Measurements Workflow.} 
\begin{table*}[h]
  \caption{\textbf{Time Consumption For Each Method Step.} We measure the average time consumption for each part of the algorithm. For SAM2, mask generation, and Inpaint Anything, we measure the average time per image $\bar t_{img}$, as it scales not linearly with the icon count. For the classification and non-maximum suppression, we measure the average time per icon, as the number of icons varies between images. Additionally, for the classification only templates are compared that have an correlation higher than 90\% of the highest correlation between template and the current icon proposal.} %
\fontsize{8pt}{8pt}\selectfont
  \centering
  \begin{tabular}{c c}
    \toprule
    \textbf{Method} & \textbf{$\bar t_{img}$ in ms} \\
    \toprule
    SAM2.1 (large) & 4,007.6 \\ %
    SAM3 & 640.8 \\
    \midrule
    Mask Generation & 676.5 \\ %
    \midrule    
    Inpaint Anything & 5,383.0 \\ %

    \bottomrule
  \end{tabular}
  \hspace{30pt}
\begin{tabular}{c c}
\toprule
    \textbf{Method} & \textbf{$\bar t_{icon}$ in ms} \\
    \toprule
    Color Histogram Comparison & 5.8   \\  %
    \midrule
    Classification LPIPS & 8.4  \\  %
    Classification CLIP & 7.6  \\ %
    \midrule
    Non-maximum-suppression & 0.018 \\  %
    \bottomrule
\end{tabular}
  
  \label{tab:time_comparison}
\end{table*}
We measure the mean time consumed by each part of the method and show the results in \cref{tab:time_comparison}. 
In the first group of method steps, the time is measured per image, in the second group per icon proposal. 
In our measurements, we ignore the time needed for loading and saving the images. 

The time $\bar t_{img}$ for the first group of steps, which are the SAM segmentation, the mask generation, and the Inpaint Anything model, is given per image, as it is independent of the number of icons per image. 
For the segmentation by SAM2.1, the consumed time depends on the number of prompts, which is in the form of a grid where objects are segmented for every given point. 
The more points are given as prompts, the longer the evaluation of the prompts takes. 

The time for the second group, namely the classification steps and the non-maximum suppression, is measured per proposed icon and then reduced to the mean.
These parts of the method are dependent on the number of icons in the images, as they have to be done for each icon individually.
This also means that the later steps are called less frequently as icon proposals are already rejected in earlier steps. 
Additionally, the runtime depends to some degree on the number of templates given, as the color histograms are compared to each template, and the image features are also compared to all templates with a similar color histogram.
The total time can be further reduced by running some parts of the method in parallel for the icons or multiple images, for example, the histogram comparison or the classification.

\noindent\textbf{Quality Comparisons.}
\begin{table*}[h]
   \caption{\textbf{Quantitative Results Dataset A.} We compare the precision, recall, and number of misclassifications across the different methods. Our method is evaluated once with cosine comparison between the CLIP features and once with LPIPS between the candidate, generated with the SAM3 model/large SAM2.1 model, and the templates. Furthermore, we evaluate both variants after inpainting with InpaintAnything(IA) the text that covers the icon proposals before classification. The presented results are evaluated with the testing set from dataset \textit{A}.}
\fontsize{8pt}{8pt}\selectfont
   \centering
   \begin{tabular}{c@{\hskip 0in}|@{\hskip 0in}c}
        \begin{tabular}{c}
        \toprule
        \\
        \toprule
        \multirow{7}*{\makecell{no add. \\ training \\ (Ours)}} \\
        \\
        \vspace{5.02pt}
        \\
        \\
        \vspace{5.02pt}
        \\
        \\
        \midrule
        \multirow{2}*{finetuned}\\
        \\
        \bottomrule
        \end{tabular}
        & 
    \begin{tabular}{c c c c c c}
     \toprule
     \textbf{Method} & \textbf{ Precision $\uparrow$ } &\textbf{ Recall $\uparrow$ } & \textbf{ Misclassifications $\downarrow$ } \\
     \toprule
     SAM2.1+CLIP & 0.9724 & 0.9640 & 0.0276  \\ %
     SAM2.1+LPIPS & 0.9900 & 0.9876 & 0.0100 \\ %
     \midrule
     SAM2.1+IA+CLIP & 0.9768 & 0.9711 & 0.0232 \\  %
     SAM2.1+IA+LPIPS & 0.9958 & 0.9940 & 0.0042 \\ %
     \midrule
     SAM3+LPIPS & 0.9875 & 0.9869 & 0.0125 \\ %
     SAM3+IA+LPIPS  & 0.9942 & 0.9937 & 0.0058 \\ %
     \midrule
     YOLOv8 & 0.9991 & 0.9996 & 0.0009  \\
     YOLOv11 & 0.9986 & 0.9994 & 0.0013 \\
     \bottomrule
   \end{tabular}
   \end{tabular}
   \label{tab:quality_comparison_methods_A}
 \end{table*}
 \begin{table*}
   \caption{\textbf{Quantitative Results Dataset B.} We compare the precision, recall, and number of misclassifications across the different methods with and without inpainting with InpaintAnything (IA). The presented results are evaluated with the testing set from dataset \textit{B} and in case of SAM2.1 with the large model.}
   \label{tab:quality_comparison_methods_B}
 \fontsize{8pt}{8pt}\selectfont
   \centering
   \begin{tabular}{c@{\hskip 0in}|@{\hskip 0in}c}
        \begin{tabular}{c}
        \toprule
        \\
        \toprule
        \multirow{7}*{\makecell{no add. \\ training \\ (Ours)}} \\
        \\
        \vspace{5.02pt}
        \\
        \\
        \vspace{5.02pt}
        \\
        \\
        \midrule
        \multirow{2}*{finetuned}\\
        \\
        \bottomrule
        \end{tabular}
        & \begin{tabular}{c c c c c c}
         \toprule
         \textbf{Method} & \textbf{ Precision $\uparrow$ } & \textbf{ Recall $\uparrow$ } & \textbf{ Misclassifications $\downarrow$ } \\
         \toprule
         SAM2.1+CLIP & 0.9676 & 0.9686 & 0.0324 \\  %
         SAM2.1+LPIPS & 0.9863 & 0.9844 & 0.0122 \\ %
         \midrule 
         SAM2.1+IA+CLIP & 0.9840 & 0.9794 & 0.0148 \\ %
         SAM2.1+IA+LPIPS & 0.9948 & 0.9953 & 0.0038 \\ %
         \midrule 
         SAM3+LPIPS & 0.9894 & 0.9869 & 0.0099 \\ %
         SAM3+IA+LPIPS  & 0.9972 & 0.9975 & 0.0022 \\ %
         \midrule
         YOLO8 & 0.9993 & 0.9999 & 0.0007 \\
         YOLO11 & 0.9995 & 1.0000 & 0.0005 \\
         \bottomrule
       \end{tabular}
   \end{tabular}
 \end{table*}
We evaluate precision and recall along with the percentage of misclassifications of detected icons.
The precision is computed as the number of correctly classified predicted icons divided by the total number of predictions, and the recall as the number of correctly classified ground truth icons divided by the total number of ground truth icons.
Both metrics are used in their original meaning of binary metrics, and not in a multi class context.
We define misclassifications as the ratio between detected icons, with the wrong icon class predicted, and the total number of predictions. 

We evaluate our method using a correlation threshold of 0.5 for both methods.
As a metric threshold for the feature comparison we use 0.7 when using LPIPS, and 0.85 when using CLIP. 
A description how we arrived at these thresholds is given in the Supplementary Material with additional precision and recall curves.
The threshold for applying non-maximum-suppression is set to 10\% coverage of one icon proposal by another one. 

As a comparison to our method, we finetuned the YOLOv8 and YOLOv11 models using the ultralytics training code~\cite{Ultralytics_YOLO} on our training sets for 40 epochs. 
The results are shown in \cref{tab:quality_comparison_methods_A} for dataset \textit{A} and \cref{tab:quality_comparison_methods_B} for dataset \textit{B}. 
In both cases, the YOLO models detect almost every icon from the ground truth, leading to a precision and recall of about 99.9\%. 
The recall and precision of our approach without inpainting evaluated with LPIPS as similarity metric is between 98.44\% and 99.0\%. 
With the use of inpainting, the results improve to values between 99.4\% and 99.75\% for precision and recall. 
Some qualitative results of the method are given in the Supplementary Material.

\section{Discussion}
From the evaluation, one can see that with a precision of around 99\% our method is already close to state-of-the-art learning-based object detection methods, such as YOLOv8, and YOLOv11, for detecting icons in navigation maps. 
In contrast to these methods, however, our method has the advantage that it does not have to be trained on a dataset, but instead works directly based on a single template image. 
Furthermore, we increase the accuracy by applying inpainting to the text in the images, leading to a precision and recall of around 99.6\%, which is even closer to the YOLO results. 

One of the main benefits of our method is that we do not need to create a dataset to train a model, which would be very time consuming.
These advantages are especially prominent if the design of the images changes frequently, as for learning-based methods, the dataset and model have to be adjusted and retrained every time, while our method can be directly applied.
Additionally, creating a training dataset requires a renderer that is already correct. However, for our particular use case, this is the component to be tested by the method. 

\noindent\textbf{Limitations.} Although our approach shows good performance for the presented use case, it will not be suitable for every use case of classical object detection, as it cannot learn unseen representations of an object. 
The optional use of inpainting, which leads to better accuracy, comes with a time consumption for masking and inpainting that is quite high compared to the other parts of the method. 

\noindent\textbf{Future Work.} 
There are two routes of future research to reduce this time consumption. 
The first is to improve the speed of inpainting models. 
The other one is to improve the classification of only partially visible icons, e.g., due to text. 
Although the accuracy is already very promising without inpainting, using it further increases the accuracy. 
If this accuracy boost can be achieved by making the classification robust to occlusion, the inpainting step can be skipped. 
This would result in a shorter runtime while having the same accuracy.

\section{Conclusion}
We presented a modern approach to template matching for the use case of detecting and classifying icons in automotive navigation maps and user interfaces. 
For our methods, we use the SAM2.1~\cite{ravi2024sam2segmentimages} and SAM3~\cite{carion2025sam3segmentconcepts} foundation models to segment all objects from the input image and then classify all icon proposals by comparing visual features from publicly available pre-trained networks. 
We evaluate the use of CLIP~\cite{radford2021learning_clip} and LPIPS~\cite{lpips} features for the classification and compare them to the object detection models YOLOv8 and YOLOv11 provided by ultralytics~\cite{Ultralytics_YOLO}.
While this approach already leads to good results, we further increase the accuracy of our classification by applying inpainting on the image to remove text that covers icons that we like to detect with automatically generated masks. 
Using that, the accuracy is only slightly worse than object detection methods, with no need to create a dataset or to train a model.
This is one core benefit of our algorithm, as it is directly applicable, requiring only the template designs that are needed for rendering the images anyway. 
In contrast to standard template matching, our method is scale invariant, robust to partial occlusions and does not have to re-run on each scale that has to be considered. %

\section*{Acknowledgements}
This work was funded by the Bavarian Transformation and Research Foundation (Project 1616-24 3DAutoQVIS). The authors gratefully acknowledge the scientific support and HPC resources provided by the Erlangen National High Performance Computing Center (NHR@FAU) of the Friedrich-Alexander-Universität Erlangen-Nürnberg (FAU) under the NHR project b264dc 3DAutoQVIS. NHR funding is provided by federal and Bavarian state authorities.

\bibliographystyle{splncs04}
\bibliography{main}

\input{supplementary}

\end{document}

%% file: supplementary.tex
\clearpage
\setcounter{page}{1}
\setcounter{section}{0}
\def\thesection{\Alph{section}}

\section{Additional Qualitative Results}
\begin{figure}[h]
     \includegraphics[width=\linewidth]{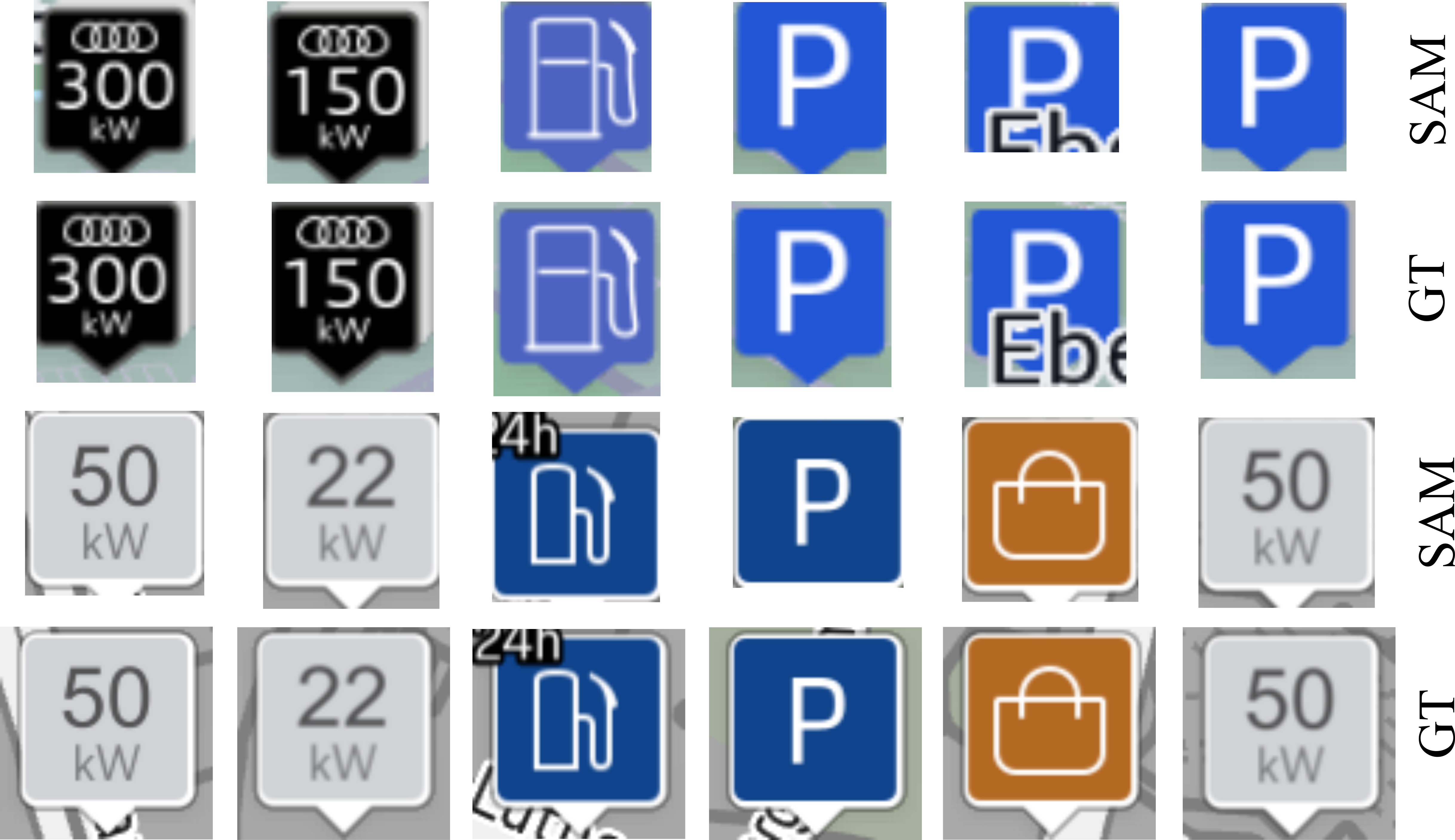}
    \caption{\textbf{Icon Bounding Boxes in Comparison.} The first two image rows show the SAM segments and ground truth bounding boxes of the icons of dataset \textit{A}, the third and fourth row of the icons of dataset \textit{B}.}
    \label{fig:sup_bbox_comparison}
\end{figure}

\begin{figure}
    \includegraphics[width=0.32\linewidth]{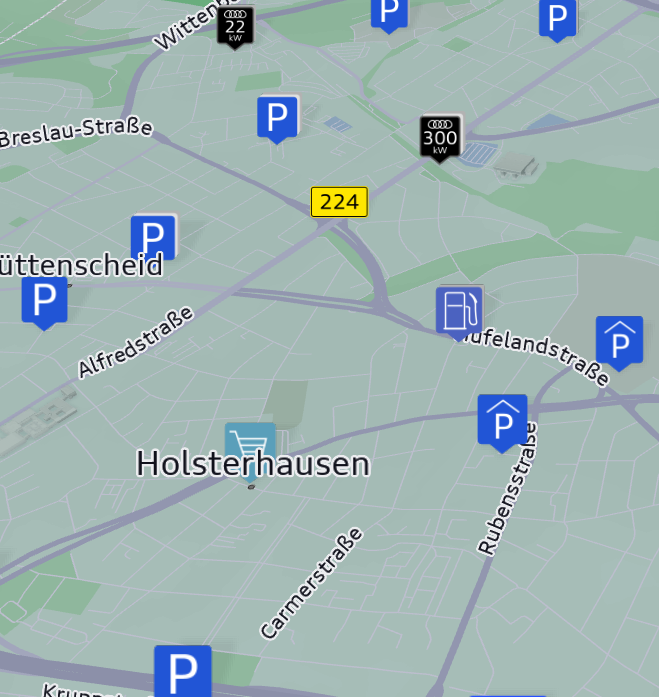}
    \includegraphics[width=0.32\linewidth]{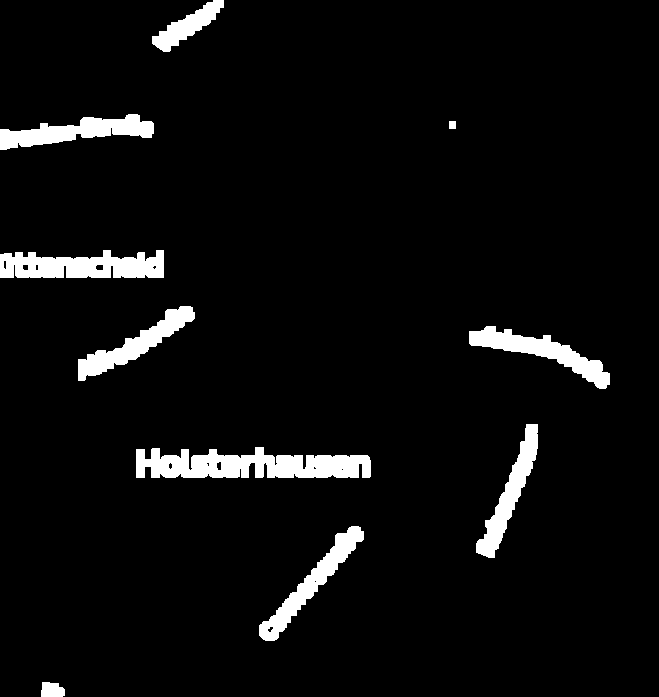}
    \includegraphics[width=0.32\linewidth]{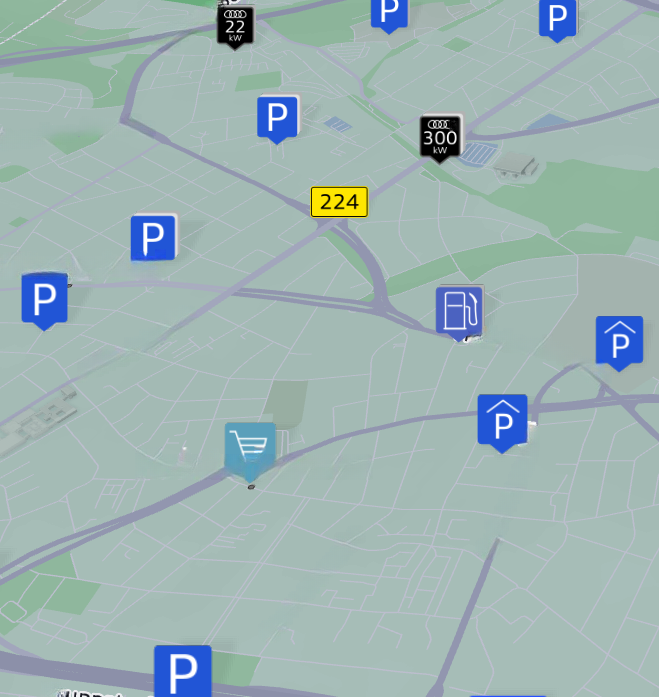}
    \caption{\textbf{Text Inpainting Results.} The images are from left to right the original image, the computed text segmentation masks, and the inpainted image. }
    \label{fig:sup_inpainted_images}
\end{figure}

\begin{figure}
     \includegraphics[width=\linewidth]{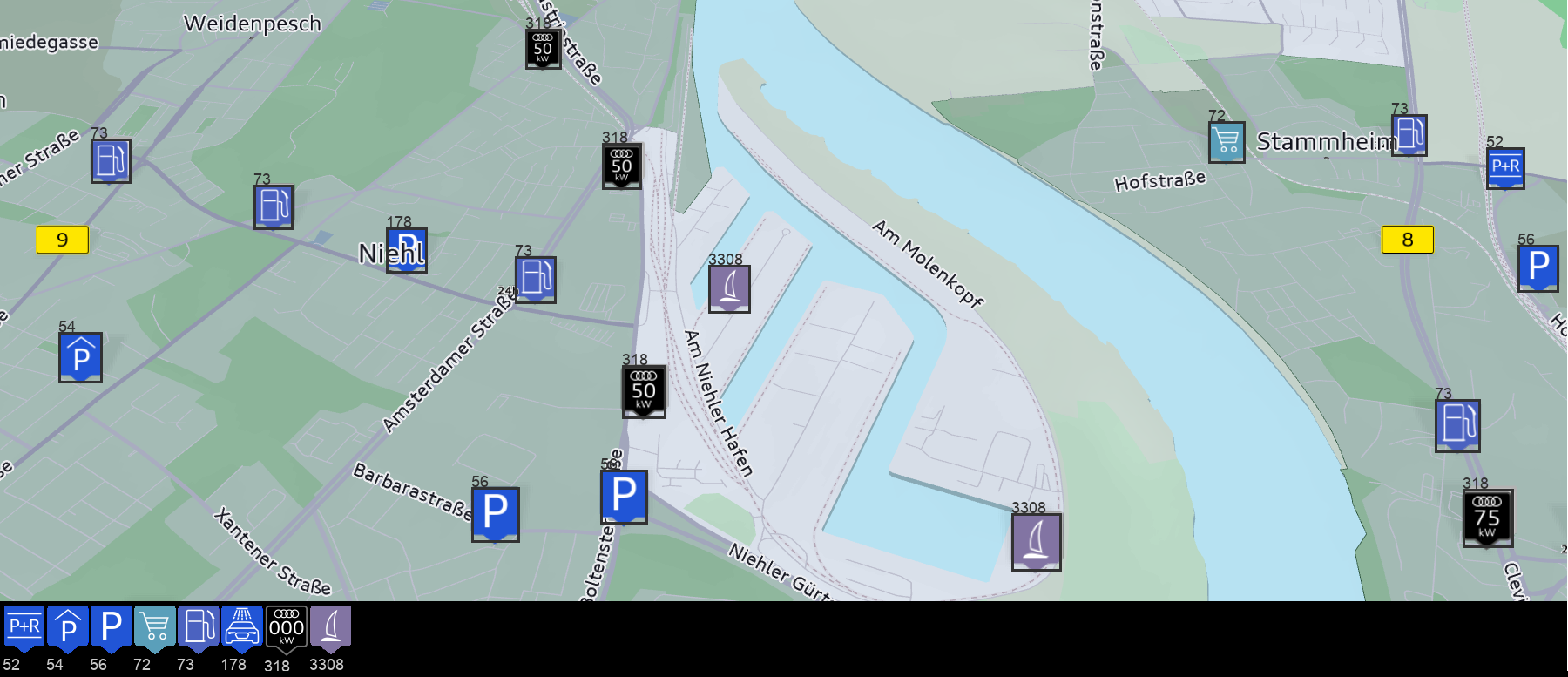}
    \caption{\textbf{Annotated Result Image.} The image contains annotations for all detected icons. Each detected icon class is shown in the lower part of the image. The image was classified by using the LPIPS variant of our method. }
    \label{fig:sup_image_annotated}
\end{figure}

\begin{figure}\begin{subfigure}{\textwidth}
    \includegraphics[width=0.32\linewidth]{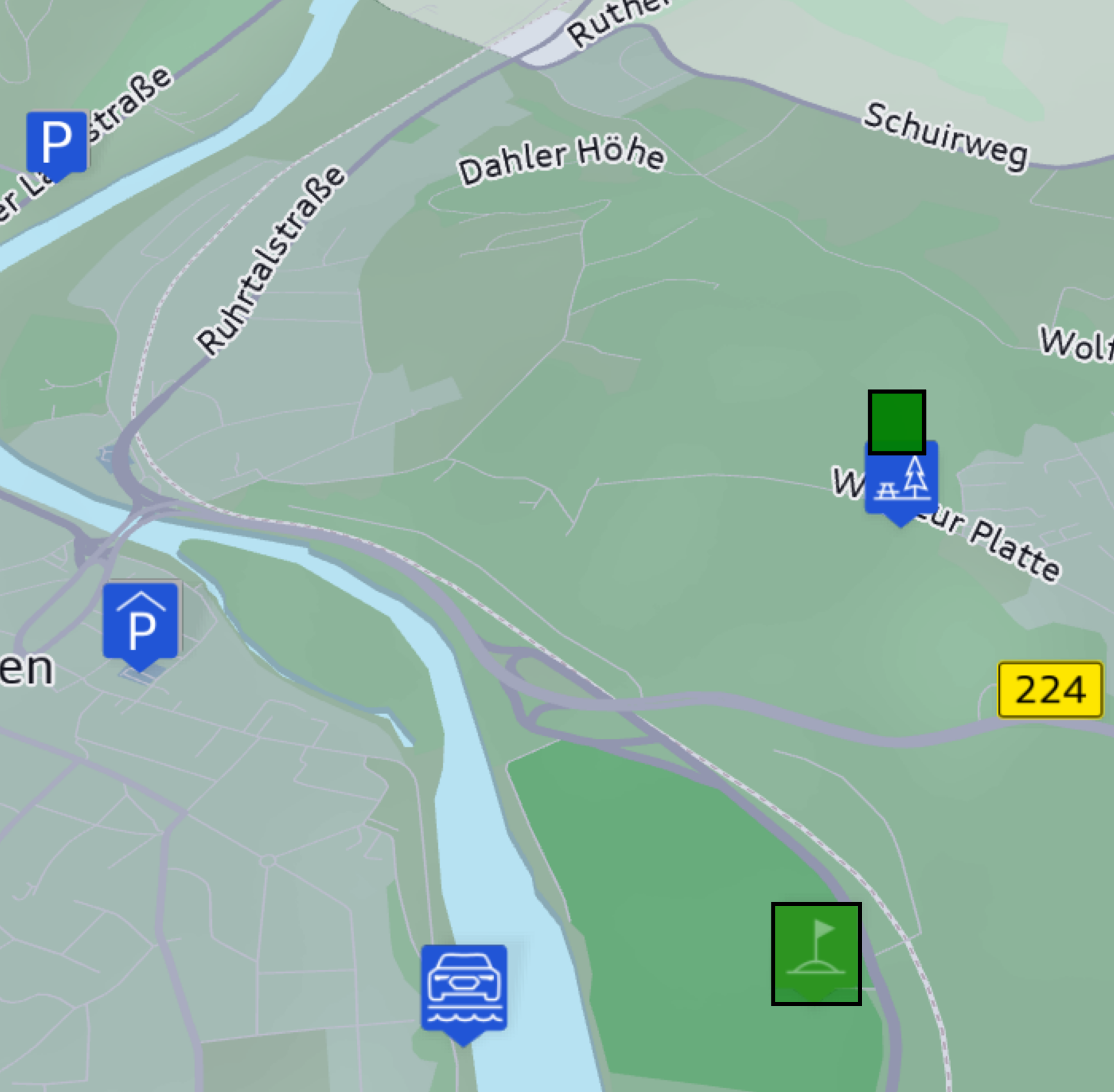}
    \includegraphics[width=0.32\linewidth]{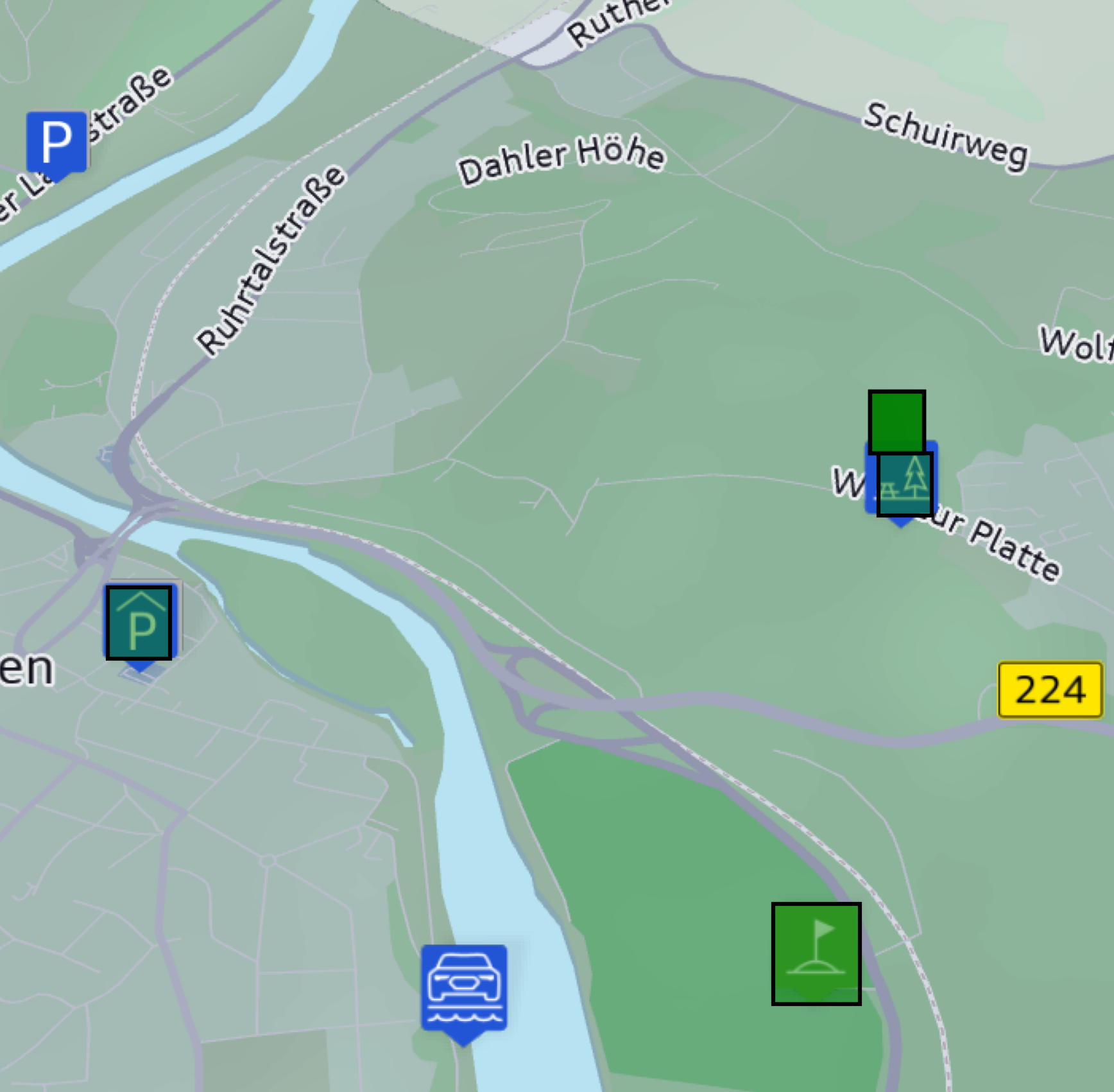}
    \includegraphics[width=0.32\linewidth]{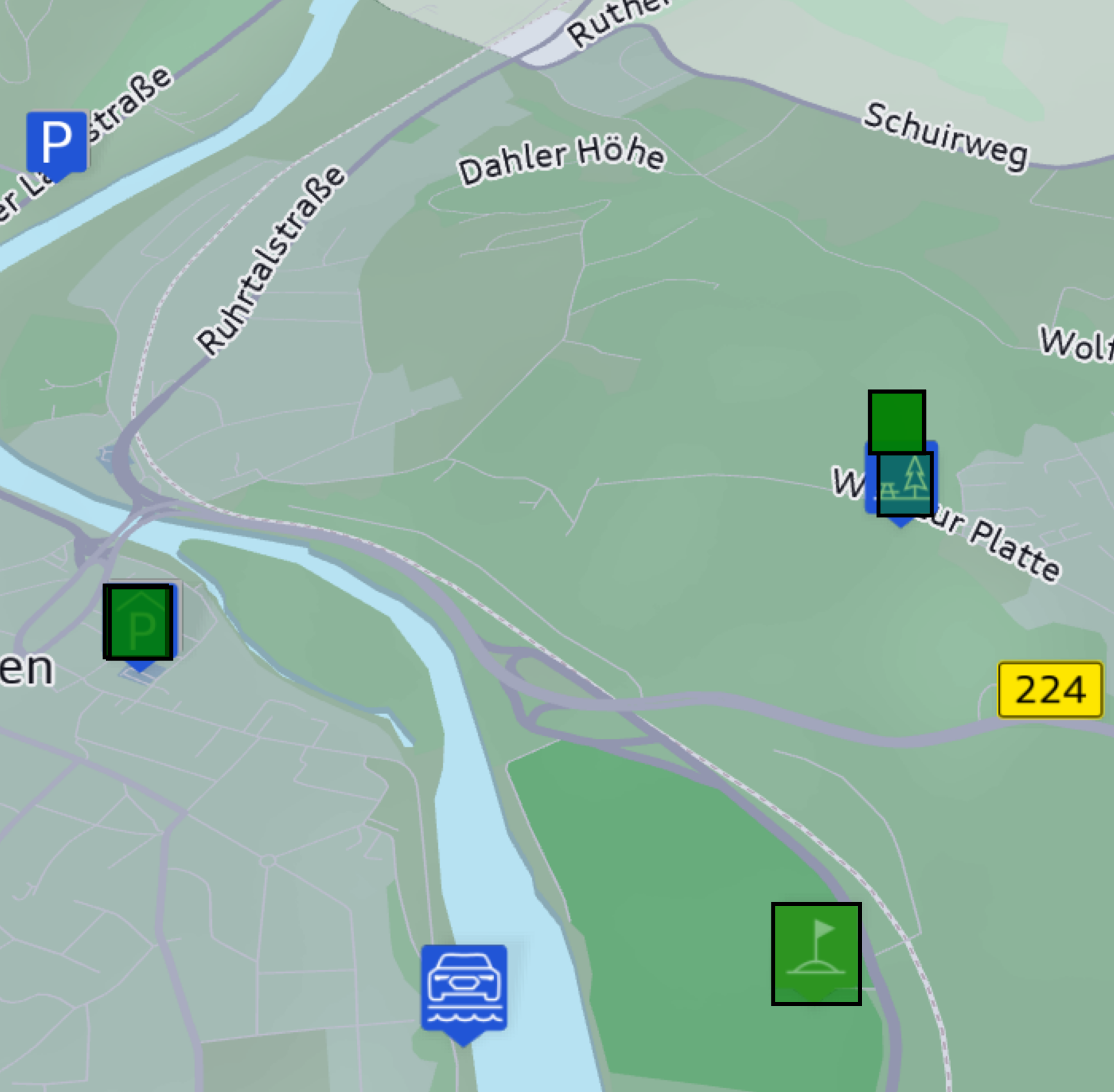}
     \caption{\textbf{Results With Varying Metric Threshold.} From left to right the metric threshold is set to 0.3, 0.35, and 0.4 with a constant color threshold of 30.}
\end{subfigure}
\begin{subfigure}{\textwidth}
    \includegraphics[width=0.32\linewidth]{images/example_images/template_matching/sqdiff/thresh_0.35_color_thresh_30.png}
    \includegraphics[width=0.32\linewidth]{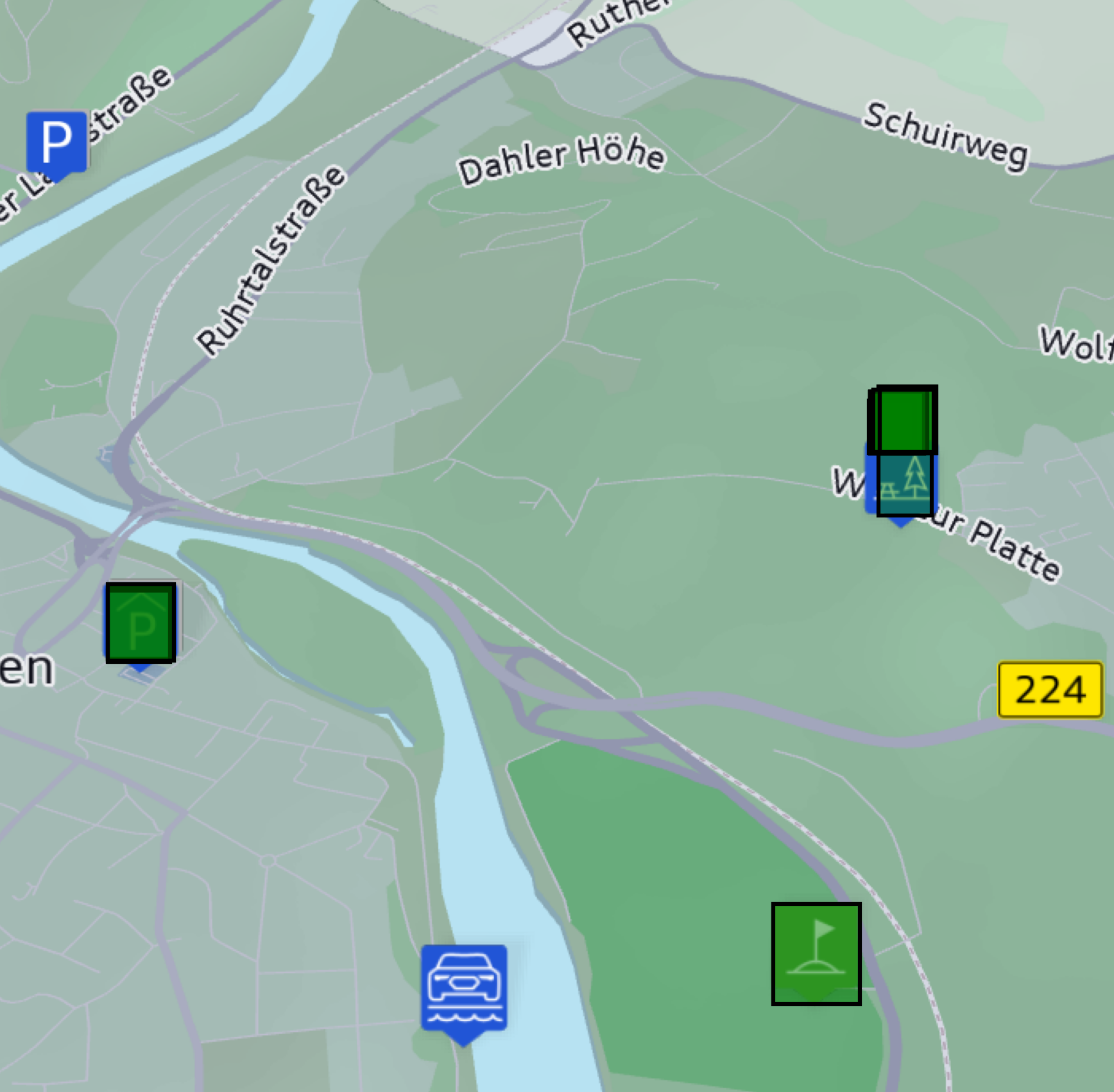}
    \includegraphics[width=0.32\linewidth]{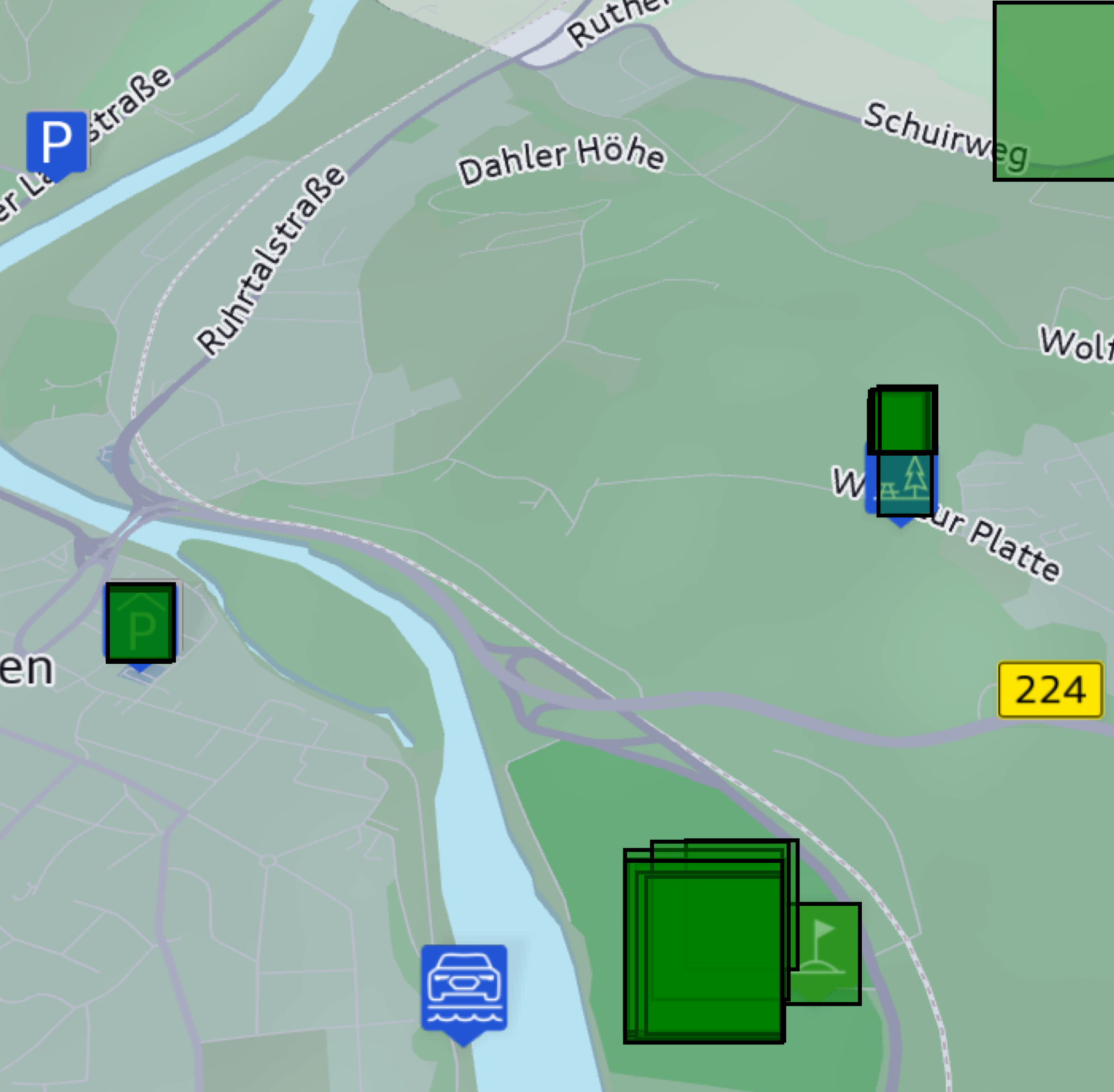}
    \caption{\textbf{Results With Varying Color Threshold.} From left to right the color threshold is set to 30, 40, and 50 with a constant metric threshold of 0.35.}
\end{subfigure}
    
    \caption{\textbf{Additional Results Template Matching.} The images show results for the template matching with all templates from the dataset with different thresholds. The templates are scaled between 50\% and 200\%. All results are gathered with the normalized squared difference as a metric. Therefore, lower metric values are better.}
    \label{fig:sup_template_matching}
\end{figure}

\begin{figure}\begin{subfigure}{\textwidth}
    \includegraphics[width=0.32\linewidth]{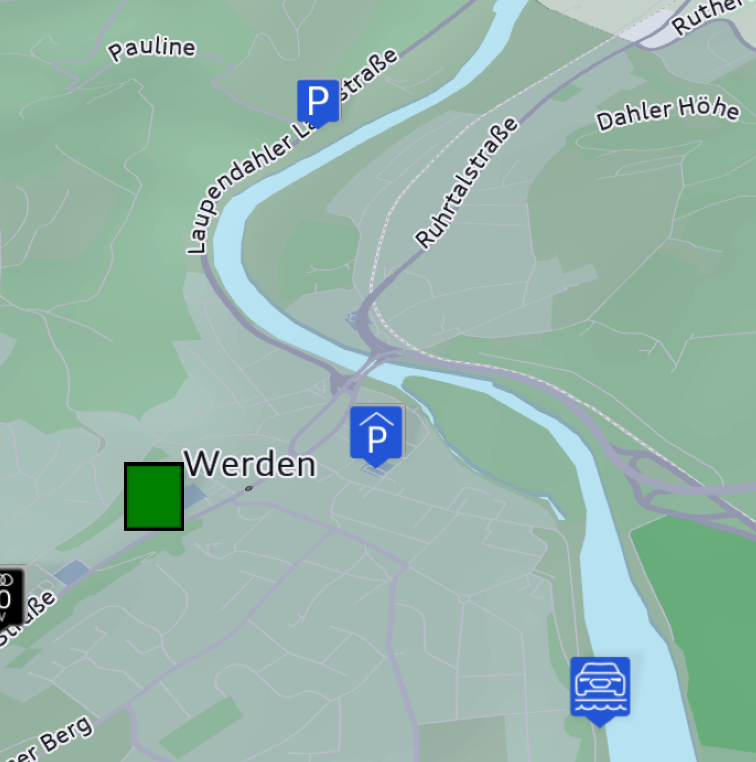}
    \includegraphics[width=0.32\linewidth]{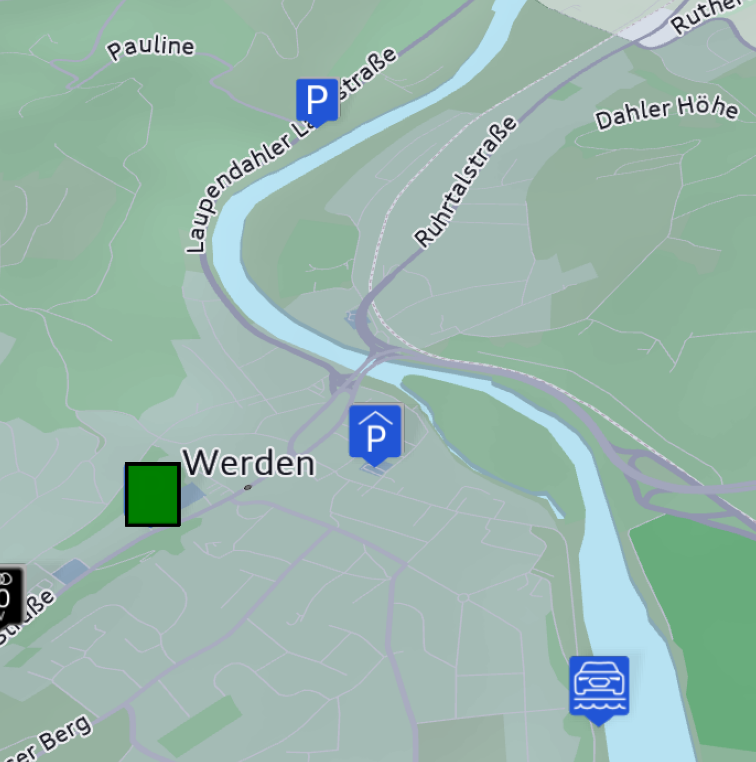}
    \includegraphics[width=0.32\linewidth]{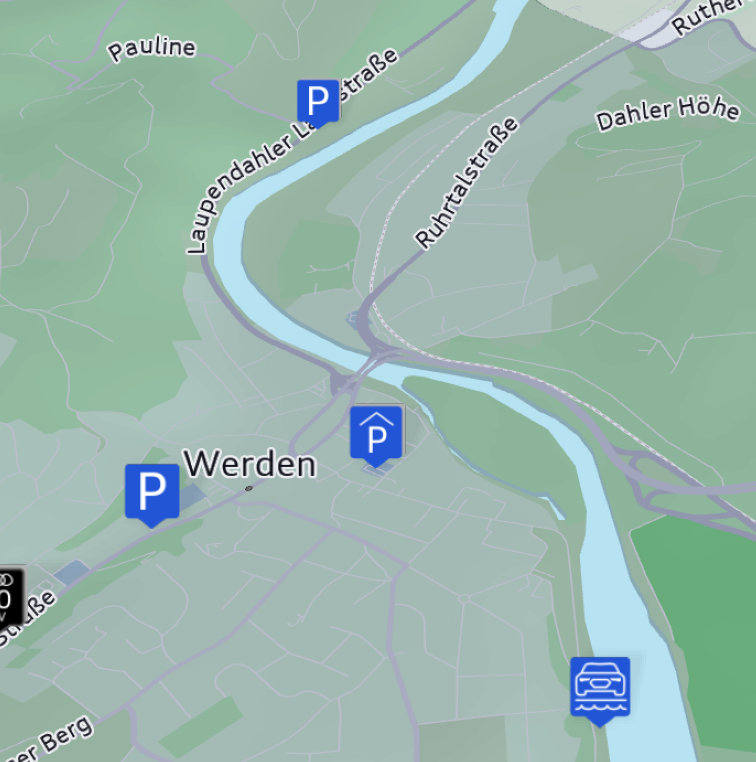}
     \caption{\textbf{Results For Parking Lot Icon.} The searched icon is the parking lot icon which is present in the image twice. From left to right the metric threshold is set to 0.83, 0.85, and 0.87 with a constant color threshold of 50.}
\end{subfigure}
\begin{subfigure}{\textwidth}
    \includegraphics[width=0.32\linewidth]{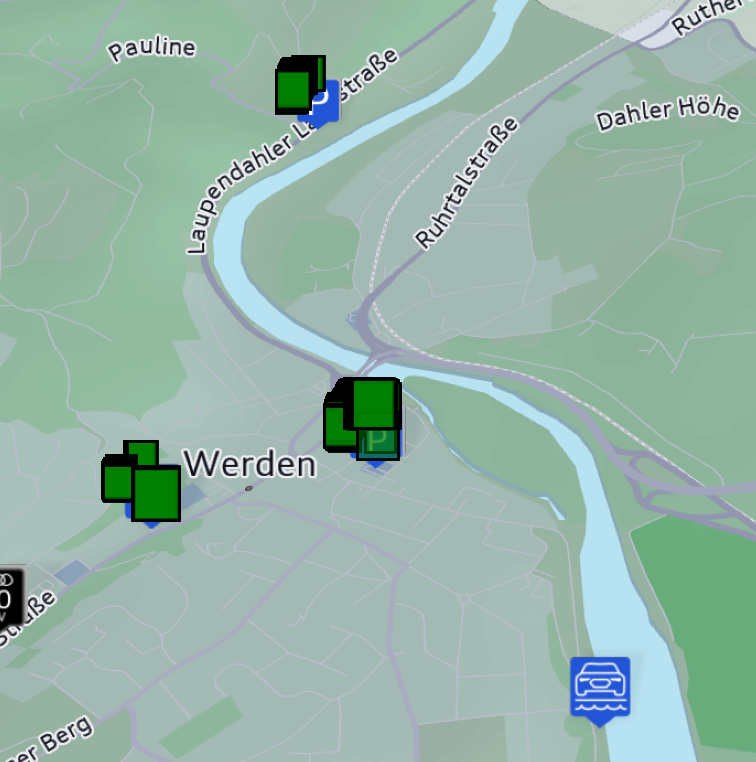}
    \includegraphics[width=0.32\linewidth]{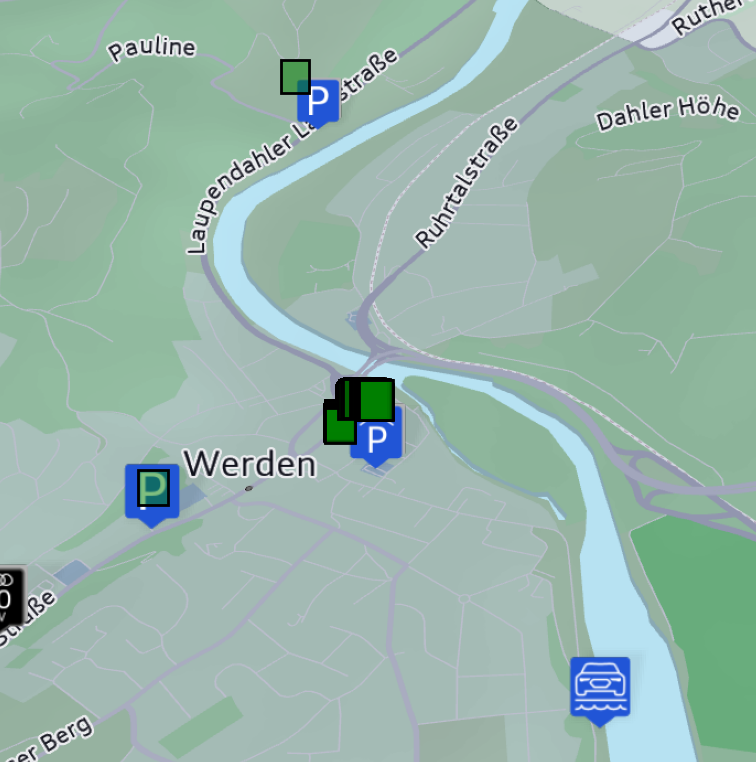}
    \includegraphics[width=0.32\linewidth]{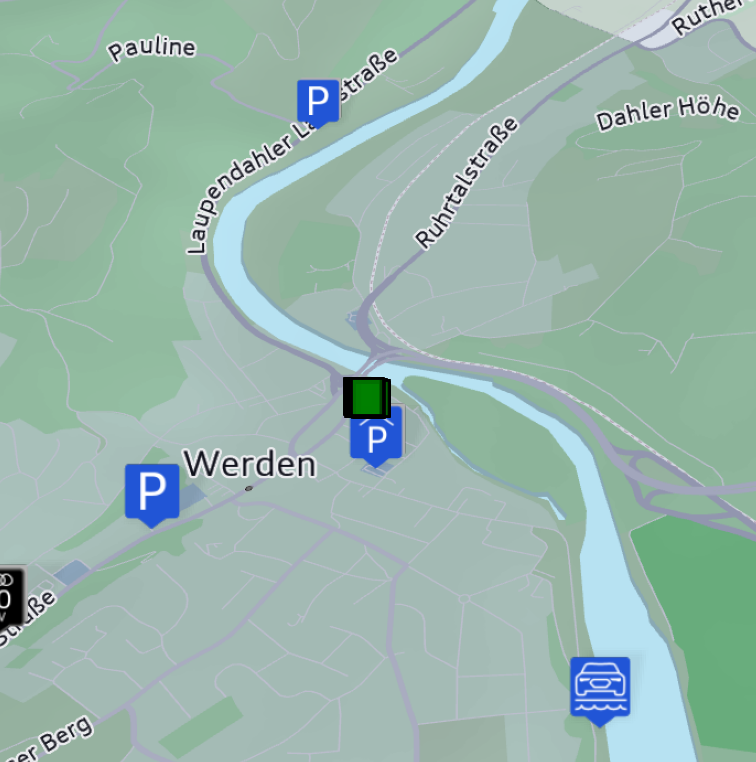}
    \caption{\textbf{Results For a Not Present Icon.} The searched icon is a gas station icon which is not present in the image. From left to right the metric threshold is set to 0.83, 0.85, and 0.87 with a constant color threshold of 50.}
\end{subfigure}
    
    \caption{\textbf{Additional Results Template Matching.} The images show results for the template matching with one template each from the dataset with different thresholds. The templates are scaled between 50\% and 150\%. All results are gathered with normalized cross correlation as a metric. Therefore, higher metric values are better.}
    \label{fig:sup_template_matching_2}
\end{figure}

\begin{figure*}
   \includegraphics[width=0.49\linewidth]{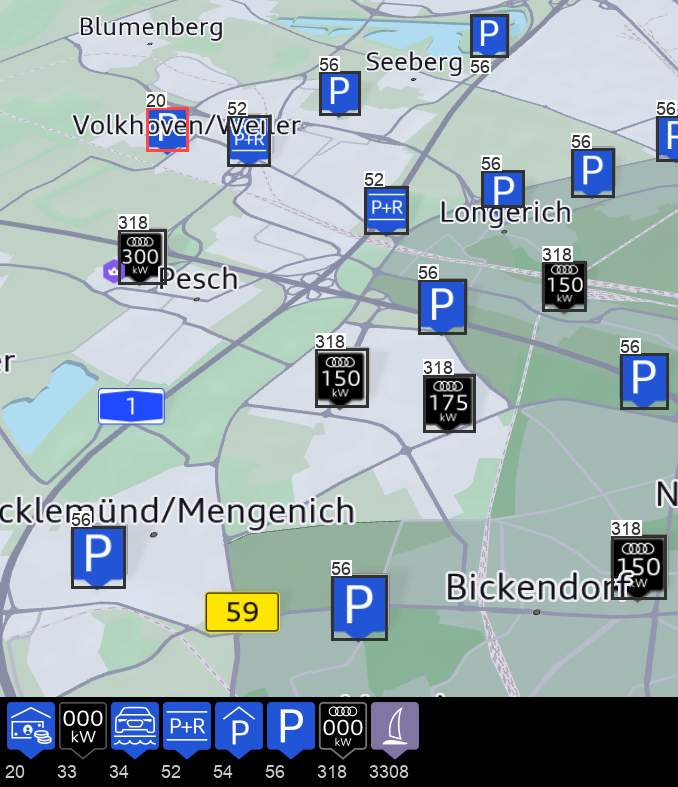}
   \includegraphics[width=0.49\linewidth]{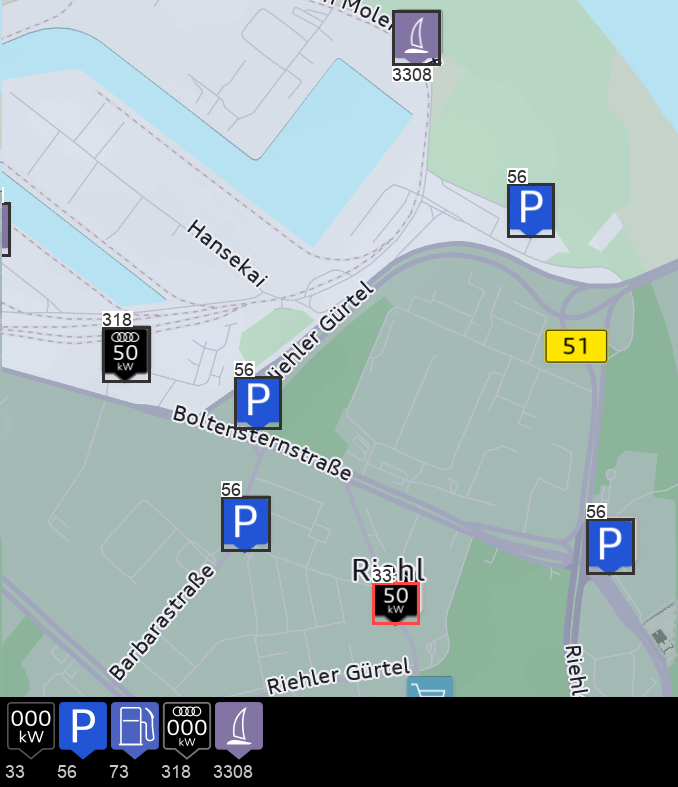}
    \caption{\textbf{Error Case Visualization.} Some error cases are depicted with red bounding boxes. The corresponding templates are annotated in the bottom line and each detected icon is labeled with the predicted class id. The results are from the not inpainted images, while using LPIPS as metric.}
    \label{fig:sup_error_visualization}
\end{figure*}

\textbf{Image Segmentation and Classification.} The show the difference between some ground truth annotations, and the corresponding detected bounding boxes in Fig.~\ref{fig:sup_bbox_comparison}. We also show a complete annotation of an image in Fig.~\ref{fig:sup_image_annotated}. Some visualizations of not detected and mismatched icons are shown in Fig.~\ref{fig:sup_error_visualization}. \\
\textbf{Inpainted Images.} We provide an example for an inpainted image together with the corresponding text segmentation mask computed as described in Sec. ~\ref{sec:methods} of the paper in Fig.~\ref{fig:sup_inpainted_images} and compare it to the original image. \\
\textbf{Template Matching.} In Fig.~\ref{fig:sup_template_matching} we provide some additional results from the template matching algorithm. We show results for the squared difference, with different thresholds for the metric value and the color similarity. Additionally, Fig.~\ref{fig:sup_template_matching_2} shows results for template matching with normalized cross correlation with when searching for individual icons. \\

\begin{figure}
     \includegraphics[width=0.7\linewidth, center]{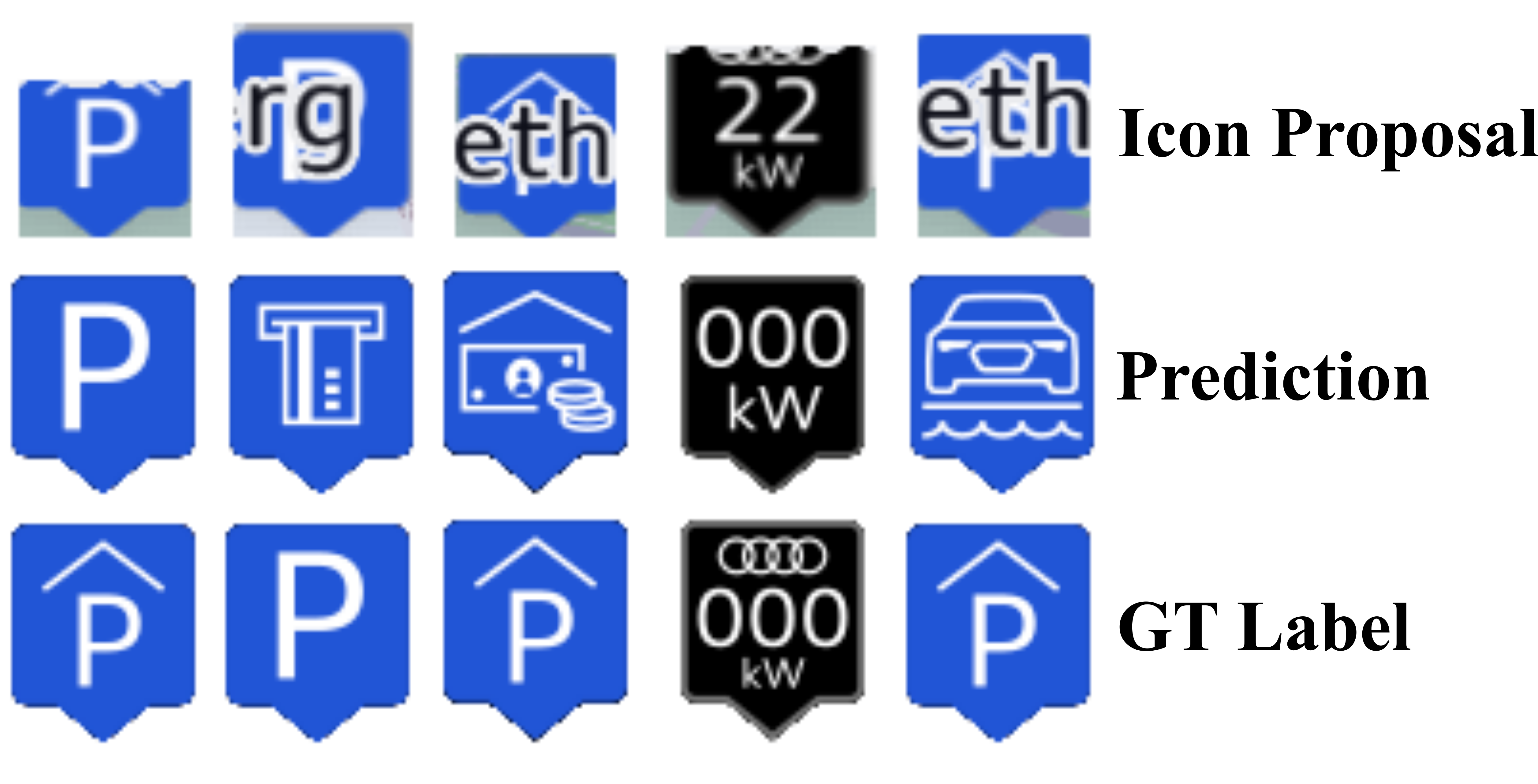}
    \caption{\textbf{Failure cases.} The first row contains the icon proposals by the SAM model. The second and third row, the predictions and ground truth labels.}
    \label{fig:failure_cases}
\end{figure}

\section{Histogram Correlation Comparison}
\begin{figure*}
    \includegraphics[width=0.49\linewidth]{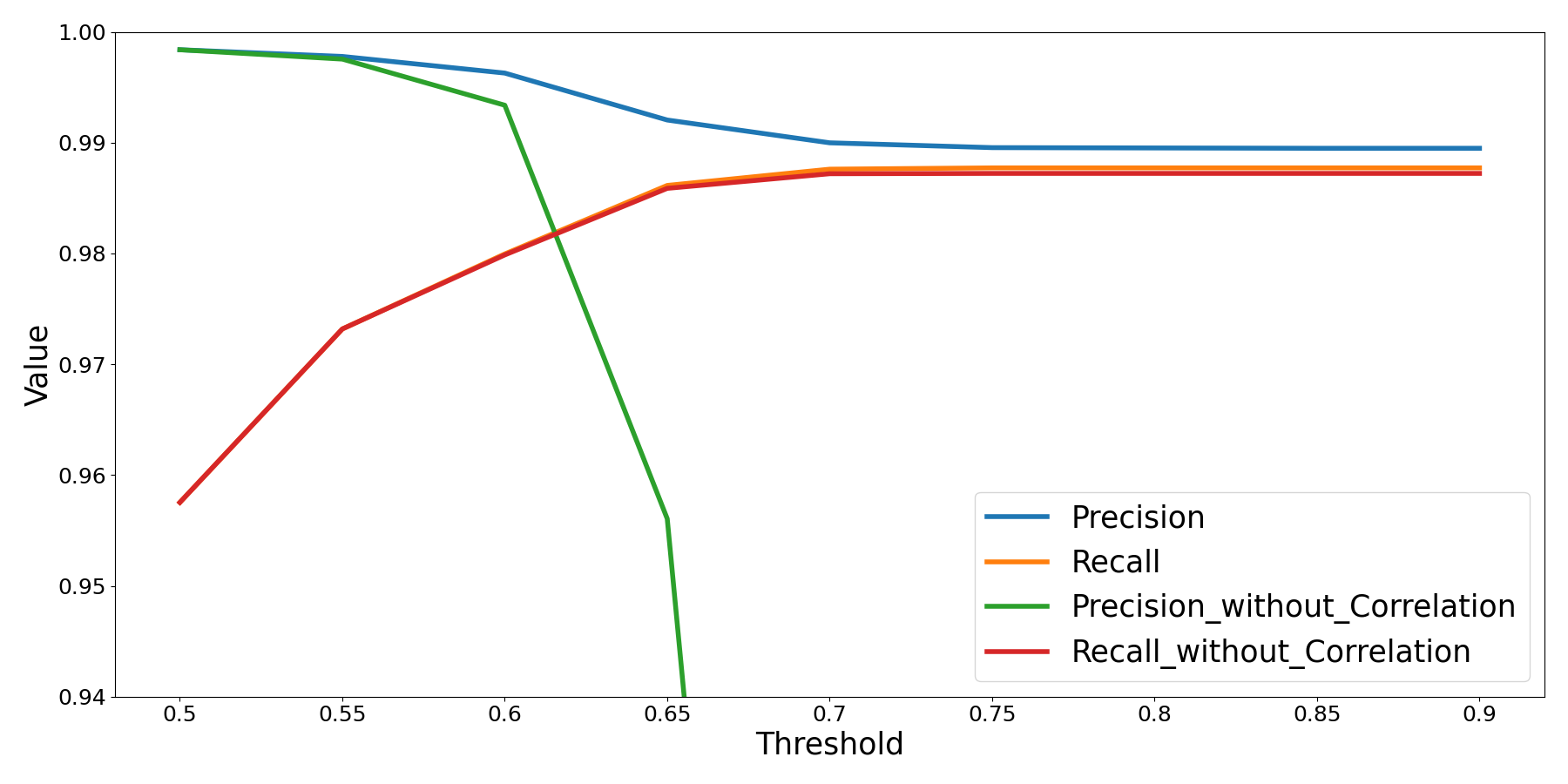}
    \includegraphics[width=0.49\linewidth]{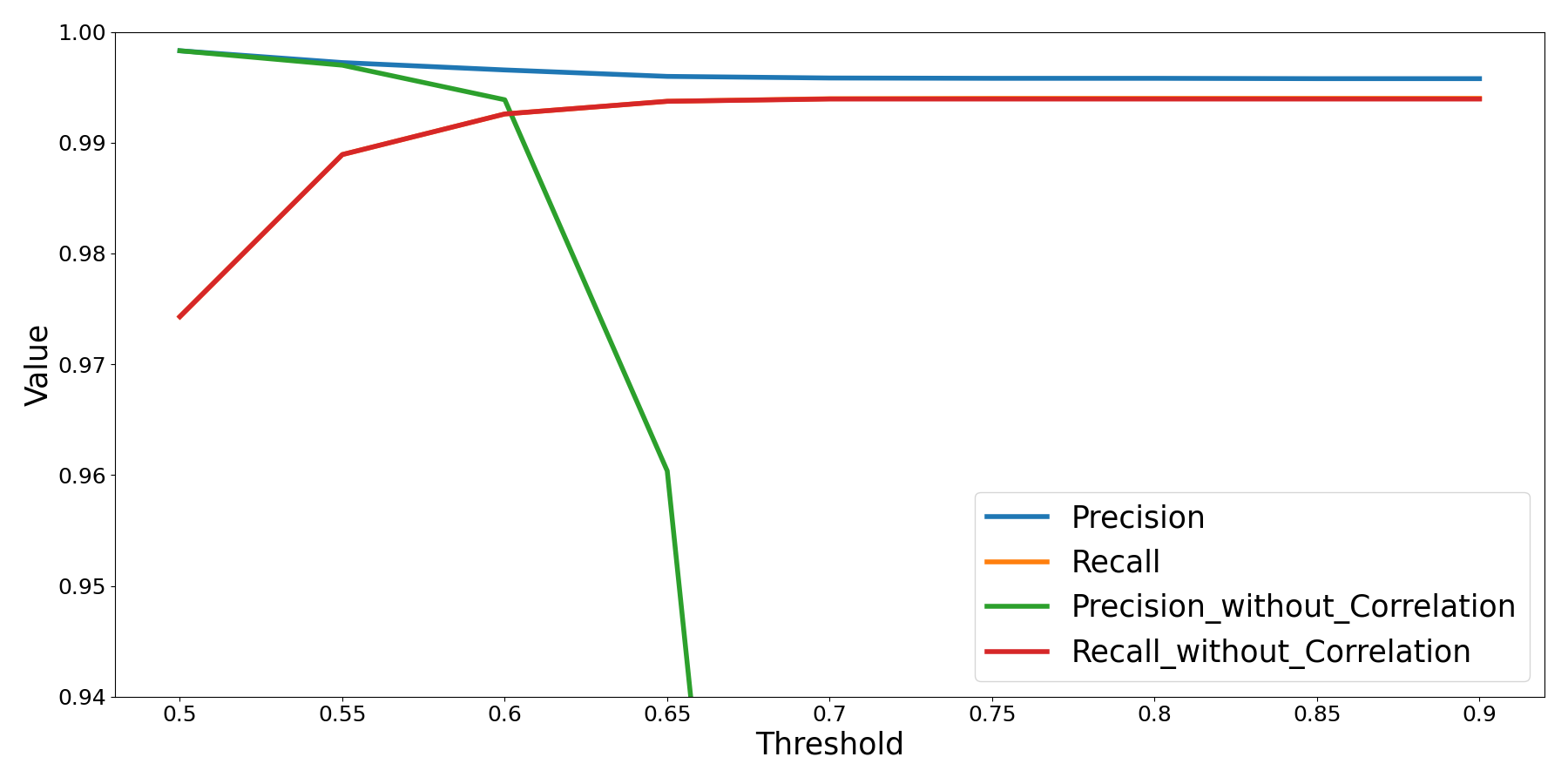}
    \caption{\textbf{Precision And Recall Using LPIPS With And Without Using Histogram Correlation.} The images show the precision and recall while changing the threshold using the LPIPS method once with applying the histogram correlation and once without. The left image shows the results without inpainting, the right one with inpainting.}
    \label{fig:sup_lpips_precision_recall_histComp}
\end{figure*}

\begin{figure}
    \centering
    \includegraphics[width=0.7\linewidth]{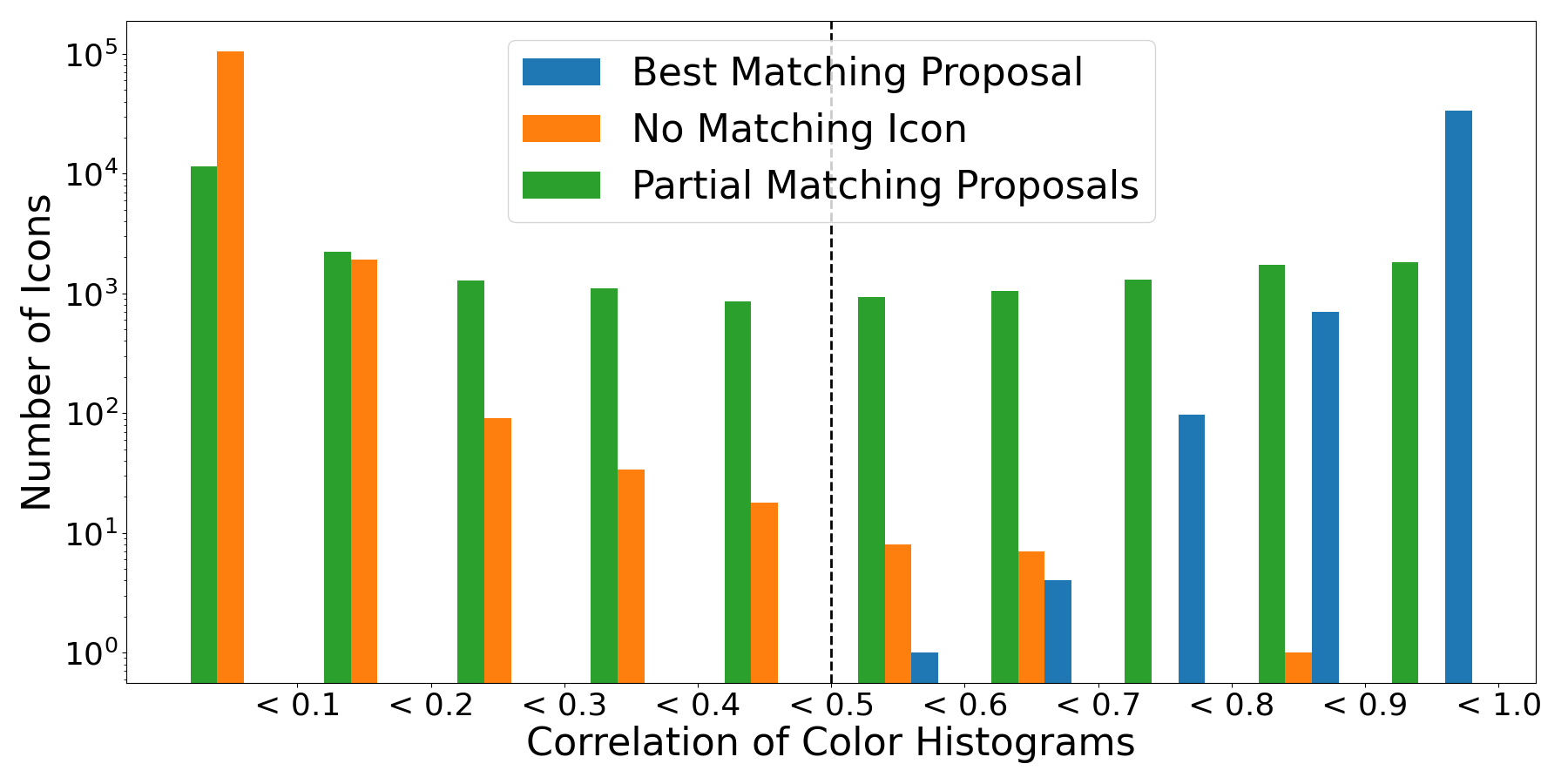}
    \caption{\textbf{Comparisons Between the Color Histogram Correlation of Icons, Partial Icons, and Non-icons}. The best correlation values for proposals without intersections with icons, and the correlation values for each icon with the best matching proposal are counted, the same is done for partial matches between icon proposals and icons. The correlation values are grouped into bins with a width of 0.1 and the number of icons are scaled logarithmically. As none of the best matches to icons has a correlation below 0.5, this part can be filtered by applying a correlation threshold (dotted line), reducing the number of icon proposals. } 
    \label{fig:correlation_plot_with_partial}
\end{figure}

\textbf{Time Comparison.} In \cref{tab:time_comparison_histogram_correlation} we evaluate the time taken to classify one icon proposal with and without optimizations. One optimization is the filtering of icons that are less than 1/4 or more than 2 times the size of the given templates. The other optimization is to only use templates for the classification that have a higher correlation than 90\% of the highest correlation between the icon proposal and the templates. \\
\begin{table}[h]
  \caption{\textbf{Time Comparison For Classification With And Without Optimization}. We evaluated the mean time consumption per icon proposal with different optimizations. One optimization is the filtering of icons that are less than 1/4 or more than 2 times the size of the given templates (Area Comp.). The other is the filtering of templates that have a correlation below 90\% of the highest correlation between the icon proposal and a template.} %
\fontsize{8pt}{8pt}\selectfont
  \centering
  \begin{tabular}[width=0.9\linewidth]{c c c c}
    \toprule
    \textbf{Metric} & \textbf{ No Optimization } & \textbf{ Area Comp.} & \textbf{ Area Comp.\& Correlation Comp.} \\
    \toprule
    LPIPS & 45.8 ms & 22.0 ms & 8.4 ms  \\
    CLIP  & 16.2 ms & 10.9 ms & 7.6 ms \\ %
    \bottomrule
  \end{tabular}
  \label{tab:time_comparison_histogram_correlation}
\end{table}
\textbf{Quality Comparisons.} In Fig.~\ref{fig:sup_lpips_precision_recall_histComp} we compare the precision and recall of our method while using the color histogram comparison and without using it. For both variants the classification is evaluated on the icon proposals by SAM2.1. By additionally using the histogram correlation a higher threshold can be selected for the LPIPS metric, which improves the achieved recall and reduces the requirement to tune the threshold to a minimum. \\
In ~\ref{fig:correlation_plot} we presented the comparison between icon proposals without having a corresponding icon and the best icon proposal. In addition to these categories there are also the icon proposals which cover the icons partially but not fully, leading to a varying highest correlation values in that category, mostly depending on how much of the icon is visible. We present this category in Fig.~\ref{fig:correlation_plot_with_partial}. 
\section{Parameter Selection}
\begin{figure*}
    \includegraphics[width=0.49\linewidth]{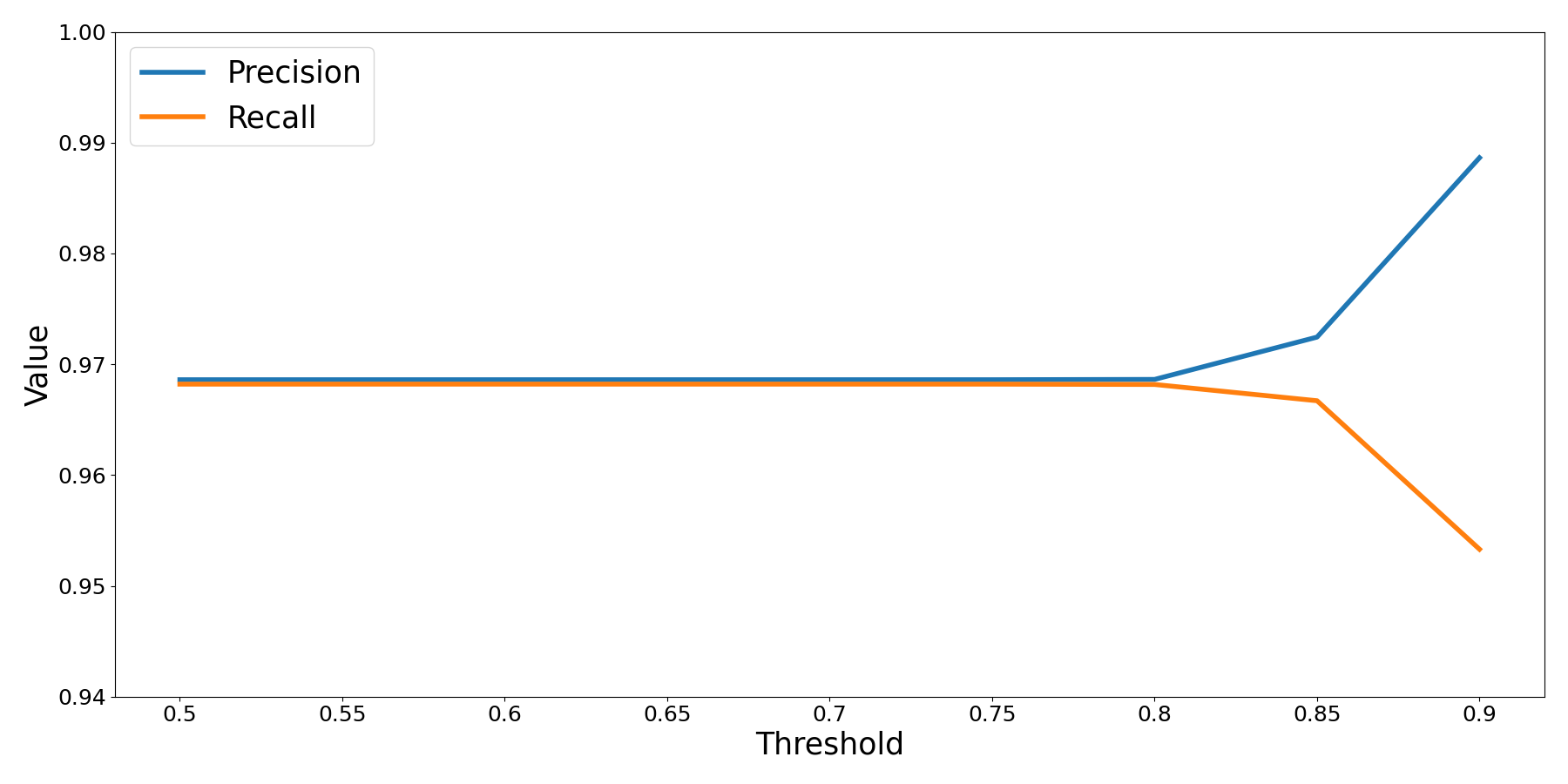}
    \includegraphics[width=0.49\linewidth]{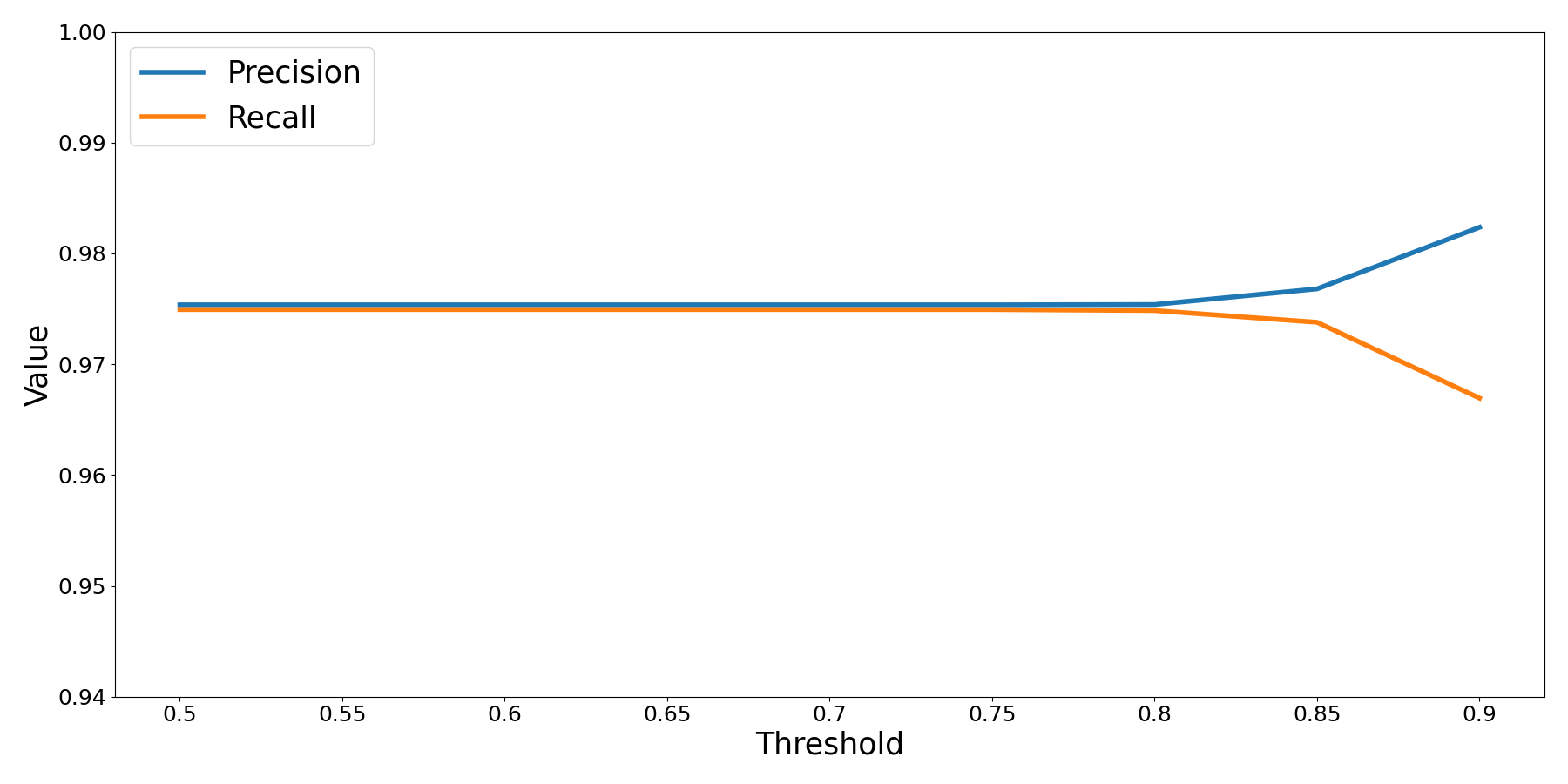}
    \caption{\textbf{Precision And Recall Using CLIP.} The images show the precision and recall while changing the threshold using the CLIP method. The left image shows the results without inpainting, the right one with inpainting.}
    \label{fig:sup_clip_precision_recall}
\end{figure*}

\begin{figure*}
    \includegraphics[width=0.49\linewidth]{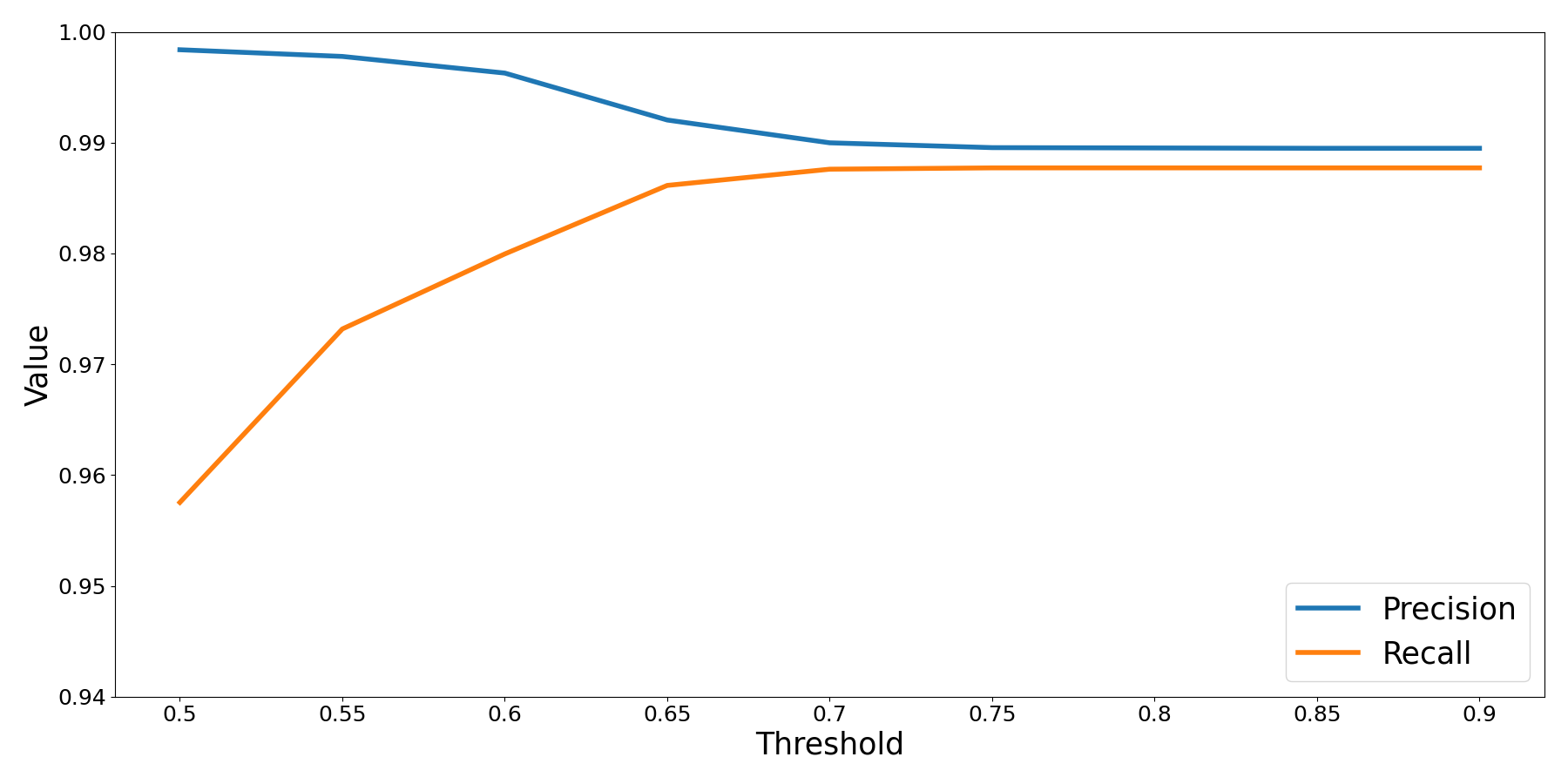}
    \includegraphics[width=0.49\linewidth]{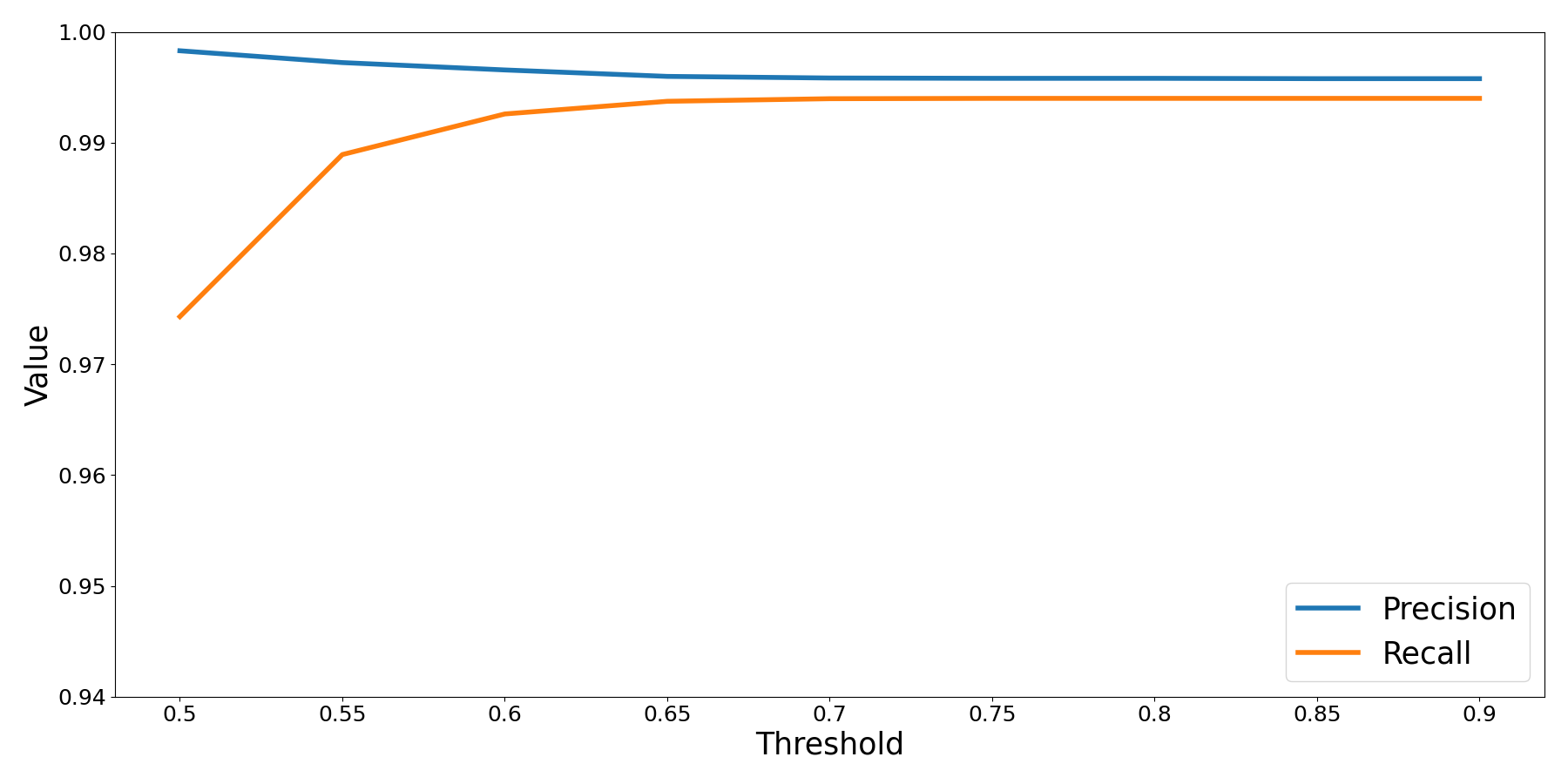}
    \caption{\textbf{Precision And Recall Using LPIPS.} The images show the precision and recall while changing the threshold using the LPIPS method. The left image shows the results without inpainting, the right one with inpainting.}
    \label{fig:sup_lpips_precision_recall}
\end{figure*}

\textbf{Metric Threshold.}
An analysis of the precision and recall of our method while changing the threshold can be found in Fig.~\ref{fig:sup_clip_precision_recall} for the CLIP metric and in Fig.~\ref{fig:sup_lpips_precision_recall} for the LPIPS metric. Both are evaluated with non-maximum suppression applied afterwards. For LPIPS the values for precision and recall do not change anymore for a threshold over 0.7, for CLIP for the thresholds below 0.85. 
This is because the remaining icon proposals are parts of an icon that are filtered by non-maximum suppression afterwards. 
Therefore, for these values the threshold has not an influence anymore.
When setting the thresholds to 0.7, respectively 0.85 for CLIP, the precision is a bit higher, while the recall is almost the same. 

